 \documentclass[final,5p,times,twocolumn,authoryear]{elsarticle}
\makeatletter
\def\ps@pprintTitle{%
 \let\@oddhead\@empty
 \let\@evenhead\@empty
 \def\@oddfoot{}%
 \let\@evenfoot\@oddfoot}
\makeatother

\usepackage{amssymb}
\usepackage{lipsum}
\usepackage{framed,multirow}
\usepackage{amsmath}
\usepackage{algorithm}
\usepackage{algorithmic}

\usepackage{amssymb}
\usepackage{latexsym}

\usepackage{url}
\usepackage{xcolor, color,soul}

\usepackage{multirow}
\usepackage{multicol}
\usepackage{adjustbox}
\usepackage{lscape}
\usepackage{graphicx}			
\usepackage{subcaption}
\usepackage{longtable}
\usepackage{ltablex,booktabs}

\begin{document}

\begin{frontmatter}

\title{SimCol3D - 3D Reconstruction during Colonoscopy Challenge}
\author[1,2]{Anita {Rau}\corref{cor1}}
\cortext[cor1]{Corresponding authors:}
\ead{arau@stanford.edu}
\author[1]{Sophia {Bano}\corref{cor1}}
\ead{sophia.bano@ucl.ac.uk}
\author[1,3]{Yueming {Jin}\corref{cor1}}
\ead{ymjin@nus.edu.sg}
\author[11]{Pablo {Azagra}}
\author[11]{Javier {Morlana}}
\author[1]{Rawen {Kader}}
\author[4]{Edward {Sanderson}}
\author[4]{Bogdan J. {Matuszewski}}
\author[5]{Jae Young {Lee}}
\author[5]{Dong-Jae {Lee}}
\author[6]{Erez {Posner}}
\author[6]{Netanel {Frank}}
\author[7]{Varshini {Elangovan}}
\author[8]{Sista  {Raviteja}}
\author[9]{Zhengwen {Li}}
\author[9]{Jiquan {Liu}}
\author[3,13]{Seenivasan {Lalithkumar}}
\author[10]{Mobarakol {Islam}}
\author[3,12]{Hongliang {Ren}}
\author[1]{Laurence B. {Lovat}}
\author[11]{Jos\'e M.M. {Montiel}}
\author[1]{Danail {Stoyanov}}

\address[1]{Wellcome/EPSRC Centre for Interventional and Surgical Sciences (WEISS) and Department of Computer Science, University College London, London, 
UK}
\address[2]{Department of Biomedical Data Science, Stanford University, Stanford, California, USA}
\address[3]{National University of Singapore, Singapore}
\address[11]{University of Zaragoza, Zaragoza, Spain}
\address[4]{Computer Vision and Machine Learning (CVML) Group, University of Central Lancashire, Preston, UK}
\address[5]{Korea Advanced Institute of Science and Technology, Daejeon, Korea}
\address[6]{Intuitive Surgical, USA}
\address[7]{College of Engineering, Guindy, India}
\address[8]{Indian Institute of Technology Kharagpur, Kharagpur, India}
\address[9]{Key Laboratory for Biomedical Engineering of Ministry of Education, College of Biomedical Engineering \& Instrument Science, Zhejiang University, China}
\address[10]{Imperial College London, London, UK}
\address[12]{The Chinese University of Hong Kong, HK, China}

\begin{abstract}
Colorectal cancer is one of the most common cancers in the world. While colonoscopy is an effective screening technique, navigating an endoscope through the colon to detect polyps is challenging. A 3D map of the observed surfaces could enhance the identification of unscreened colon tissue and serve as a training platform. However, reconstructing the colon from video footage remains difficult. Learning-based approaches hold promise as robust alternatives, but necessitate extensive datasets.
Establishing a benchmark dataset, the 2022 EndoVis sub-challenge SimCol3D aimed to facilitate data-driven depth and pose prediction during colonoscopy. The challenge was hosted as part of MICCAI 2022 in Singapore. Six teams from around the world and representatives from academia and industry participated in the three sub-challenges: synthetic depth prediction, synthetic pose prediction, and real pose prediction. This paper describes the challenge, the submitted methods, and their results. We show that depth prediction from synthetic colonoscopy images is robustly solvable, while pose estimation remains an open research question.
\end{abstract}


\end{frontmatter}

\section{Introduction}
The Endoscopic Vision (EndoVis) challenges at MICCAI have been an accelerator for surgical data science for several years 
\citep{maier2017surgical,maier2020bias,maier2022surgical}. Past challenges have evaluated a range of tasks such as segmentation, image generation, or action triplet detection\footnote{\url{https://endovis.grand-challenge.org/}}. Although the applications are widely different, all challenges share a profound contribution to their respective research fields by improving data availability and bringing attention to research gaps. In the spirit of this tradition, the \textit{SimCol3D - 3D Reconstruction during Colonoscopy} challenge was born. SimCol3D is the first challenge to contribute both synthetic and real colonoscopy procedure sequences to address depth estimation and 6D pose estimation from monocular colonoscopy.

Colorectal cancer (CRC) is a leading cause of death \citep{araghi2019global}, third only to lung and breast (for female) and prostate (for male) cancer. Despite its prevalence, survival rates are high among individuals undergoing screening \citep{kaminski2010quality}. The slow progression of CRC allows for an extended window for detecting and treating pre-cancerous growths. But to be treated, such growths first need to be accurately detected---an exceedingly difficult task. Fortunately, a cohort of AI-based platforms has declared missed polyps a relic of the past \citep{puyal2022polyp,ji2021progressively,zhao2022semi,chadebecq2023artificial}, making it possible to assist clinicians in identifying polyps on the colon mucosa during colonoscopy in real-time. Yet, challenges persist, particularly in detecting polyps hidden behind folds, which constitute up to three-quarters of all missed polyps \citep{pickhardt2004location}.  Additionally, other lesions, such as dysplasia in Inflammatory Bowel Disease (IBD) patients, pose an exceptional challenge, necessitating meticulous screening of the entire colon mucosa. The quality of the screening is often quantified as the time taken to withdraw the colonoscope, a critical aspect of the procedure for lesion detection. Withdrawal time is a key surrogate marker for adenoma detection rate \citep{butterly2014serrated}, which, in turn, is associated with post-colonoscopy CRC rate \citep{corley2014adenoma}. But withdrawal time as a measure of performance has significant limitations. It measures overall time and fails to ensure sufficient attention to each colon segment. A 3D map could help provide more useful quality indicators such as withdrawal time per segment, or ratio of screened colon mucosa. 

Researchers have thus proposed to generate an on-the-fly 3D map of the colon during a colonoscopy that can flag areas of the colon that need to be re-screened for colorectal polyps. 
But providing such a map is difficult. The poor quality of real colonoscopy videos, caused by artifacts such as specularities, air bubbles, blur, saturated pixels, and lack of contrast \cite{ali2021deep}, presents a significant hurdle to feature-based methods. Repetitive textures and geometries, extreme deformation, and challenging and view-dependent lighting additionally challenge feature matching between images. 

Data-driven approaches circumvent the need for robust features and divide the task into depth prediction and pose estimation. But despite the significant progress made by deep learning in reconstructing 3D scenes \cite{zhou2017unsupervised}, the translation of such approaches to colonoscopy is limited by data availability. While cities or rooms can be scanned using lidar or infrared sensors, such scanners are not deployable within a spatially constrained colonoscope. To date, there exists no dataset containing RGB images, camera poses, and depth maps from a real colonoscopy. Attempts to work around this limitation, such as registering previously acquired computer tomography (CT) scans of the colon with images from the procedure, fail due to the immense deformation of the colon during its inspection. Similarly, the calibration of non-medical-grade structured-light sensors, electric-magnetic tracking sensors, and standard colonoscopes is exceptionally difficult, often inaccurate, and only applicable to phantoms that deviate significantly in visual and haptic characteristics from real colons. Synthetic data, though visually distinct, offers precise and abundant annotations.

Previous work leveraging synthetic depth data mostly focused on bridging the domain gap between real and synthetic data \citep{mahmood2018deep,rau2019implicit,mathew2020augmenting,cheng2021depth,itoh2021unsupervised,rodriguez2022uncertain} and employed existing depth networks. In contrast, the challenge organizers were curious to explore depth prediction without accounting for the domain shift between real and synthetic data and chose to evaluate methods directly on synthetic depth. Data-driven pose prediction had yet to be explored widely before the SimCol3D challenge, mostly due to the lack of camera pose ground truth \citep{rau2022bimodal}. We therefore provided synthetic and real pose labels to differentiate between the scenario in which pose networks can be learned in a supervised manner and a scenario where no ground truth is available.  

We believe a  mapping technology for colonoscopy to be within reach and created the SimCol3D challenge to bringing us one step closer to reliable 3D reconstruction of the colon.
 
In this paper, we
\begin{itemize}
    \item introduce the SimCol3D challenge: the first of its kind for depth and pose prediction in colonoscopy;
    \item analyze each participating group's results, identifying trends and best practices across three subtasks: synthetic depth prediction, synthetic pose estimation, and real pose estimation;
    \item establish a benchmark for future comparisons of depth and pose estimation methods in colonoscopy;
    \item introduce synthetic data based on two additional human CT scans, augmenting our existing dataset;
    \item highlight avenues for future investigation. 
\end{itemize}

\section{Related work}
To date, a profound gap exists between research efforts in depth prediction and pose estimation, both intrinsic subtasks of 3D reconstruction.   Depth prediction solves the task of regressing or classifying each pixel in an image, and such tasks are more easily learnable for neural networks if sufficient training data exists. However, understanding camera movement and its geometric implications through regression alone is a much more challenging task and remains underexplored \citep{rau2022bimodal}. Some works have thus focused on leveraging the depth and pose networks in a mutual framework. This section briefly reviews essential works in the field to give context to the participants' contributions. 

Most works on depth prediction during colonoscopy have two things in common: they are borrowed from general computer vision approaches, and they incorporate synthetic data in some way. Some notable virtually generated or phantom-based public datasets were proposed by \cite{rau2019implicit}, \cite{zhang2020template}, \cite{ozyoruk2021endoslam}, \cite{bobrow2022colonoscopy}, and \citep{rau2022bimodal}. While they were an important addition to the research community, they all consist of one anatomy only, and cannot be used to evaluate accuracy on an unseen patient.
\cite{mahmood2018deep} proposed one of the first depth networks for colonoscopy and is based mainly on convolutional neural fields proposed in \citep{liu2015learning}. The authors trained one network for depth prediction on synthetic data and used a second, independently optimized network to translate between the appearance of real and synthetic images. \cite{rau2019implicit} use the well-known pix2pix network \citep{isola2017image} to integrate the depth and domain translations networks into a single framework trained on both synthetic and real data. \cite{cheng2021depth} propose to train a well-known GAN \citep{wang2018high} on synthetic data with supervision and, in a second, independent step, train the initialized network on real images with self-supervision. \cite{mathew2020augmenting} base their method on the well-known CycleGAN network that maps virtual images to real images and vice versa. \cite{itoh2021unsupervised} also borrow the cycle-consistency losses from CycleGAN and decompose images based on a Lambertian-reflection model to train their network on synthetic and real data. \cite{rodriguez2022uncertain} based their method on MonoDepth2 \citep{godard2019digging} and trained an ensemble method with a teacher trained on synthetic data. 
Though these methods help progress the field, all of these method primarily focus on bridging the domain gap between synthetic and real images, not on improving the architectures of the respecitve depth networks. Accordingly, the evaluation protocols focused on real colonoscopy frames that are oftentime borrowed from in-house datasets. A common benchmark allowing a systematic comparison of these methods is missing.  

While methods that predict depth only largely rely on synthetic data, approaches combining depth and pose networks can directly learn from real data. \cite{bae2020deep} use sparse SfM pseudo ground truth to supervise their colon reconstruction pipeline. They reconstruct small colon sections from eight consecutive frames using the derived poses and sparse depth supervision to guide the initial U-Net based \citep{ronneberger2015u} depth estimation. \cite{ma2019real} propose a  SLAM pipeline that integrates a well-known recurrent neural net for depth and pose estimation \cite{wang2019recurrent}. \cite{freedman2020detecting} and \cite{ozyoruk2021endoslam} propose self-supervised networks based on the popular depth and pose networks \citep{gordon2019depth} and \citep{bian2019unsupervised}, respectively. All these approaches do not require synthetic data; however, they can only be as accurate as the underlying feature-based SfM reconstruction.   Additionally, these works use in-house datasets and do not provide a sufficient comparison between each other.

\section{Tasks and datasets}

\begin{figure}
    \centering
    \includegraphics[width=\linewidth]{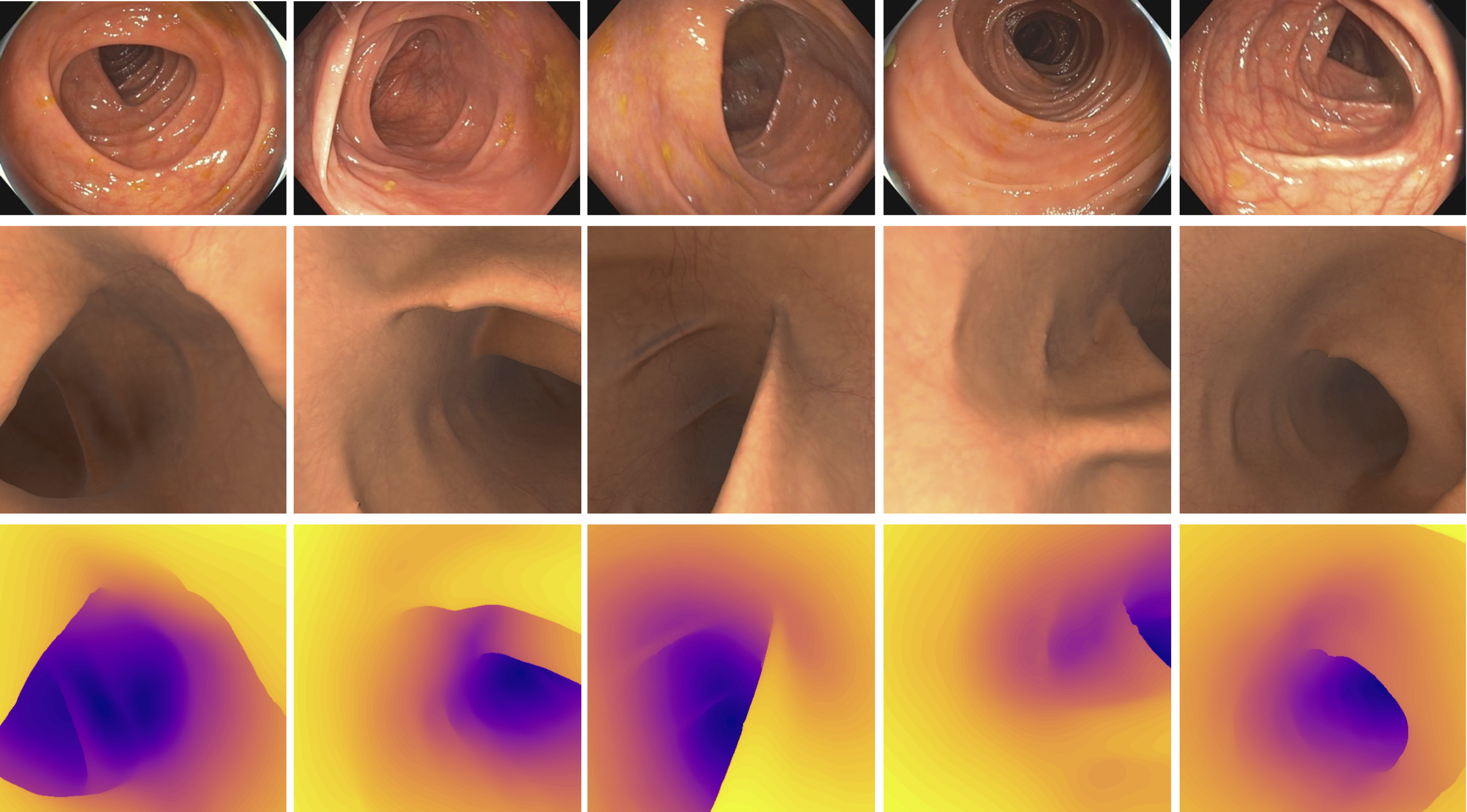}
    \caption{Overview of the real images (top), synthetic images (center), and synthetic depth maps (bottom) used in the challenge.}
    \label{fig:data_overview}
\end{figure}

\subsection{Challenge Tasks}
The SimCol3D challenge aims to facilitate depth and camera pose prediction during colonoscopy by providing a new public dataset with ground truth depths and poses for training and testing.
The challenge comprises three tasks: Task 1 invited participants to train networks to predict depth from simulated colonoscopy images. Task 2 evaluates predicted camera poses from simulated colonoscopy. 
Task 3 extends the challenge into the realm of real-world clinical practice, tasking participants with predicting poses from real colonoscopy procedures. 

\subsection{Data}
The SimCol3D challenge encompasses both synthetic and real colonoscopy sequences. Table \ref{tab:sub-datasets} provides an overview of the data used in the challenge, and Figure \ref{fig:data_overview} shows illustrates qualitative examples.

\begin{table}
    \centering
    \resizebox{\linewidth}{!}{
    \begin{tabular}{l|c|c|c}
    \hline
    \textbf{Sub-dataset	}& \textbf{\# Train traj.}& \textbf{\# Test traj.} & \textbf{	\# Images}\\
    \hline
    Synthetic colon I (Public mesh)	& 12	&3	&18k	\\
    \hline
    Synthetic colon II (Patient A)	&12&	3	&18k\\
    \hline
    Synthetic colon III (Patient B)	&0	&3	&1.8k	\\
    \hline
    Real Sequences&	59&	7&	-	\\
    \hline 
    \end{tabular}}
    \caption{Overview of the datasets in the SimCol3D challenge, indicating the number of trajectories (traj.) and images per scene. The real sequences provide videos only.}
    \label{tab:sub-datasets}
\end{table}

\subsubsection{Simulated colonoscopy data for Tasks 1 and 2}

 The synthetic data for Tasks 1 and 2 builds upon the dataset introduced in \citep{rau2022bimodal}, but expands its scope from one anatomy (Synthetic Colon I) to encompass three distinct human colons (Synthetic Colons I, II, and III). Two of the three subsets (namely I and II) contain 15 trajectories of which 12 were randomly assigned for training and three for testing. Synthetic Colon III only contains 3 trajectories for testing and no training data. This setup allows to evaluate generalizability to new anatomies.  
\begin{figure}[h]
    \centering
    \includegraphics[width=\linewidth]{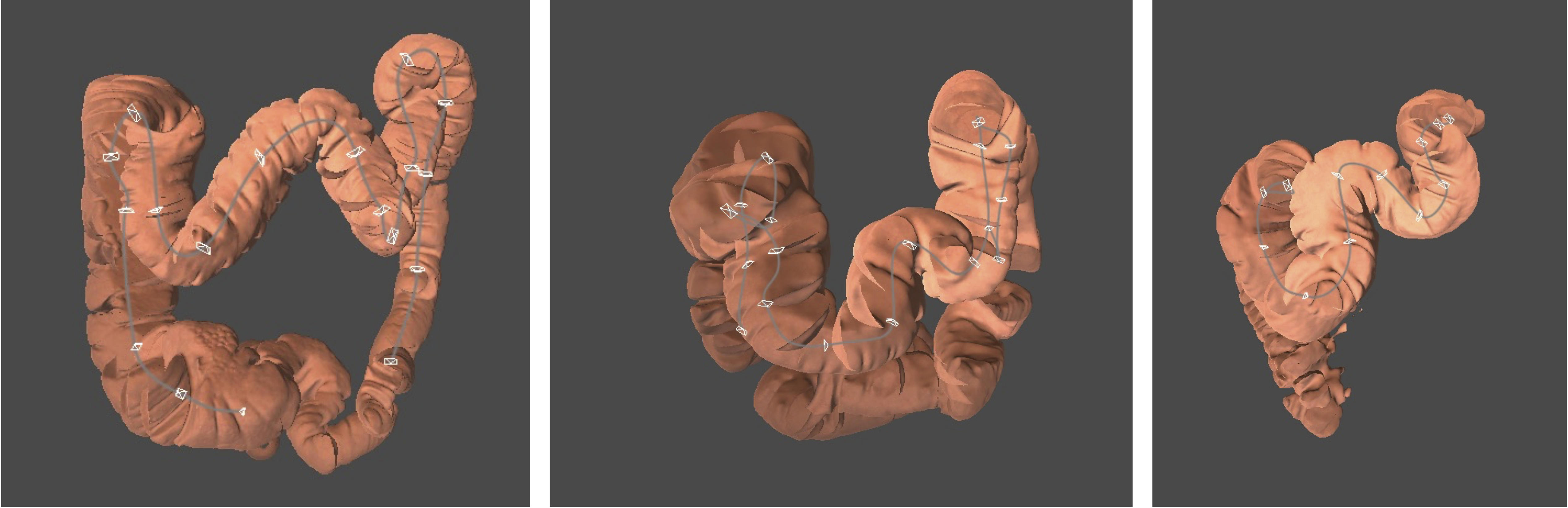}
    \caption{Synthetic Colons I, II, and III in Unity environment with camera paths along the center of the mesh. Synthetic Colons I and II include training and test trajectories. Synthetic Colon III provides test trajectories only.}
    \label{fig:enter-label}
\end{figure}
Each training trajectory contains 1201 images, ground truth depth maps, and ground truth camera poses. Each test trajectory contains either 1201 or 601 frames and their labels.  The simulated colon meshes were extracted from computer tomography scans of human colons, and the images were rendered using a Unity simulation environment \citep{rau2019implicit}. The CT scan for Synthetic Colon I is publicly available \citep{ozyoruk2021endoslam}, while the CT scans for Colons II and III were acquired at University College London Hospital. In the simulation environment, a virtual colonoscope followed a path through the center of the meshes recording rendered RGB images, depth maps, and camera poses.  This path was randomly manipulated each time a new trajectory was recorded resulting in different, random trajectories within the same anatomy.  
In total, the training data for Task 1 and Task 2 contains 14,412 frames, and the test set contains 9,009. For each frame the challenge organizers provided the corresponding:
\begin{itemize}
    \item $3 \times 3$ camera intrinsics matrix saved as \textit{txt} file. 
    \item Depth map in \textit{png} format,  including the depth value for each pixel in the corresponding RGB image. 
    \item  Absolute camera pose in the Unity coordinate frame provide in a .txt file. We represent camera pose as 7D vector $[t_x, t_y, t_z, q_x, q_y, q_z, q_w]$, where $t$ denotes the translation along the $x$-, $y$-, and $z$-axes, and $q$ denotes the rotation in quaternion representation, where $w$ denotes the scalar part, and $x$, $y$, and $z$ describe the imaginary parts.  
\end{itemize}

 More details about the data generation process and the coordinate systems used to represent the data can be found in the original publication \citep{rau2022bimodal}.  The full synthetic dataset is publicly available here: \url{https://www.ucl.ac.uk/interventional-surgical-sciences/simcol3d-data}. Participants were allowed to use additional datasets as long as they were publicly available.
\subsubsection{Real patient data for Task 3}
For Task 3, the testing data contained three patients’ anatomies with 1–3 trajectories from each and 7 in total. The real data comes from the EndoMapper dataset \citep{azagra2023endomapper}, which is a collection of complete endoscopy sequences obtained during regular medical procedures\footnote{\url{https://www.synapse.org/Synapse:syn26707219/wiki/615178}}. It includes 59 sequences with over 15 hours of video and is the first endoscopic dataset to include geometric and photometric endoscope calibration. The dataset also includes meta-data and annotations. The participants were encouraged to train their models on the EndoMapper sequences that were not in the test set. For this task, we generated COLMAP pseudo ground truth of the 7 testing sequences for method evaluation. As COLMAP is not reliable in colonoscopy (therefore the need for this challenge), two challenge organizers   and a gastroenterologist , qualitatively verified each of the generated COLMAP trajectories and sparse point clouds and chose those that were visually coherent with respect to the   direction of the movement of the endoscope   observed in the corresponding video.

\subsection{Evaluation metrics}
\subsubsection{Task 1: Depth estimation}
We utilize three standard evaluation metrics to assess the performance of the depth prediction methods. We define the per image errors as
\begin{align}\label{eq:depth metrics}
    L_{1} & = \frac{1}{D}{\sum_{d = 1,...,D}}||Y(d)-s\cdot{Y^{'}{(d)}}||_{1} \\
    L_{rel} & = \mu_d\left ( \left \Vert \frac{Y(d) - s \cdot Y'(d)}{Y(d)} \right \Vert_1 \right ) \\
    L_{RMSE} & = \sqrt{\frac{1}{D}\sum_{d = 1,...,D}(Y(d)-s\cdot Y^{'}(d))^2} 
\end{align}
where $Y$ denotes the ground truth depth map, $Y^{'}$ denotes the predicted depth map, $D$ is the number of pixels in $Y$, and $\mu_d$ represents the median calculated for all valid arguments $d$. Let
\begin{align}
    \bar{Y}_i = \frac{1}{D}{\sum_{d=1,...,D}} Y_i(d),
\end{align}
denote the mean depth over all pixels in a depth map $i$, then the scale $s$ is calculated per trajectory as 
\begin{align}
    s = \frac{\sum_{i\in I}\bar{Y}_i \cdot \bar{Y}^{'}_i}{\sum_{i\in I} \bar{Y}^{'}_i\cdot \bar{Y}^{'}_i},
\end{align}
where $I$ denotes the number of images in a trajectory.   We chose to evaluate the scaled depths, as the task of monocular depth estimation is ill-posed and networks are expected to predict depth up to scale.  
We compute the $L_{1}$ loss as the mean of the absolute differences between the ground truth depth $Y(d)$ and the predicted depth $Y^{'}(d)$ over all pixels in a depth map. As the relative loss, $L_{rel}$, is sensitive to outliers, we use the median instead of the mean over the per-pixel relative $L_{1}$ errors. Lastly, we measure the $L_{RMSE}$ as it weights outliers more heavily than the $L_{1}$ loss. The per-depth map errors are then averaged over all depth maps in a scene.

As we found all three metrics to be equally descriptive of performance, but due to their different scales not comparable, we use a point system for Task 1. We report the final score, $\sum_1$, as the sum of ranks per scene. For each of the three scenes and each of the three metrics, the winner received six points, the runner-up five points, etc. The task winners were the groups with the most points. 

\subsubsection{Task 2: Camera pose estimation on simulated data}

To evaluate the predicted camera poses, we first composite the relative poses $\Omega_i$ to produce the complete trajectory of absolute poses $P_i$. The absolute pose of a camera $\tau$ in the world space is $ P_{1}\Omega_{1}\cdots\Omega_{\tau-1}$, where each $\Omega_i$ sequentially projects the initial pose $P_1$ to the next one. As monocular video can only be interpreted up to scale, the predicted trajectory needs to be scaled using:
\begin{equation}
        s_{rel} = \frac{\Sigma_{\tau}trans(\Omega_{\tau})^T\cdot trans({\Omega'_{\tau}})}{\Sigma_{\tau}trans({\Omega'_{\tau}})^T\cdot trans({\Omega'_{\tau}})},
\end{equation}
where \textit{trans} denotes the translation of a projection matrix.
We then assess the scaled predicted trajectory's accuracy with the Absolute Translation Error ($ATE$), Relative Translation Error ($RTE$), and Rotation Error ($ROT$).
\begin{equation}\label{pose metrics}
    \begin{aligned}
        RTE & = \mu_{\tau}(||trans(\Omega_{\tau}^{-1}\Omega^{'}_{\tau})||) \\
        ATE & = \mu_{\tau}(||trans(P_{\tau})-trans(P^{'}_{\tau})||) \\
        ROT & = \mu_{\tau}(\frac{trace(Rot(\Omega_{\tau}^{-1}\Omega^{'}_{\tau}))-1}{2}\cdot\frac{180}{\pi}) \\
    \end{aligned}
\end{equation}
where $Rot$ denotes the projection rotation,  $\Omega^{'}, P^{'}$ are the scaled predicted relative and absolute poses,  and $||\cdot||$ is the two-norm. The $ATE$ measures drift and the overall consistency of a predicted trajectory. The $ROT$ measures the magnitude of the rotation errors locally.   The $RTE$ reflects both translation and rotation errors locally. To achieve a small $RTE$, the predicted relative pose $\Omega'$ must be close to the ground truth $\Omega$, so that $\Omega_{\tau}^{-1}\Omega^{'}_{\tau}$ is close to an identity matrix. This is achieved, when both $trans(\Omega')$ and $Rot(\Omega')$ are accurate.   We consider the forward direction only. Evaluating these three evaluation metrics, we obtain a comprehensive assessment of the performance of the pose prediction models. To determine the winner of Task 2, we define the task loss $\sum_2$ as the weighted average of RTEs on the three scenes, where we weight SynCol III twice to account for the increased difficulty of pose prediction on an unseen scene.
\subsubsection{Task 3: Camera pose estimation on real-world data}
We use the same evaluation metric for Task 3 as for Task 2. In particular, we determine ATE, RTE, and ROT as defined in Equation \ref{pose metrics}. However, we scale the entire trajectory based on the absolute poses to reflect that we are more interested in the global consistency in Task 3, than in local accuracy. The scaling factor in Task 3 is defined as:
\begin{equation}
        s_{abs} = \frac{\Sigma_{\tau}trans(P_{\tau})^T\cdot trans({P'_{\tau}})}{\Sigma_{\tau}trans({P'_{\tau}})^T\cdot trans({P'_{\tau}})},
\end{equation}
The task score $\sum_3$ for Task 3 has three components: ATE, RTE, and ROT averaged over all seven scenes.
\subsection{Challenge organization}
 
The challenge was a one-time event with fixed submission deadline of September 2022. In order to access the train and test data, participants had to register participation in the challenge on the challenge website\footnote{\url{https://www.synapse.org/Synapse:syn28548633/wiki/}}. The teams provided their predictions for the test sets via the challenge website based on detailed submission guidelines including docker templates and evaluation scripts that participants could use for validation \footnote{\url{https://github.com/anitarau/simcol}}. The ground truth for the test data was published after the challenge had ended. The participants were not required to publish their code, but links to the code bases of the teams that chose to are provided in Section 4. Ethics approval was not necessary for this challenge. In total, we received and approved 51 challenge registration requests and 13 team registration requests. 
 
\section{Methods for Task 1: Depth prediction from synthetic images}

\begin{table*}
\caption{Summary of the participating teams of the SimCol Challenge – Task 1}
\centering
\resizebox{\linewidth}{!}{
\begin{tabular}{l|l|c|l|l|l|l} 
\hline
\multirow{1}{*}{\textbf{Team name}} & \multicolumn{1}{c|}{\textbf{Algorithm}} & \multicolumn{1}{c|}{\textbf{T/C}} &\multicolumn{1}{c|}{\textbf{Loss function}} & \multicolumn{1}{c|}{\textbf{Preprocessing~}} & \multicolumn{1}{c|}{\textbf{Data augmentation}} & \multicolumn{1}{c}{\textbf{Post-processing}}          \\ 

\hline
\begin{tabular}[l]{@{}l@{}} CVML  \end{tabular}       &\begin{tabular}[l]{@{}l@{}}FCBFormer adaptation \\ \citep{sanderson2022fcn} \end{tabular} &T\&C&MSE &\begin{tabular}[l]{@{}l@{}}Alpha channel removed, pixel intensity \\normalization, resize, depth scaling \end{tabular} &Horizontal \& vertical flips &None
\\ 
\hline
\begin{tabular}[l]{@{}l@{}} EndoAI  \end{tabular}       &\begin{tabular}[l]{@{}l@{}}GLPDepth \\ \citep{kim2022global} \end{tabular} & T\&C &SILog &Normalization, Horizontal flip & Vertical cut depth &\begin{tabular}[l]{@{}l@{}}Conv2D-ReLU-Conv2D \\ block to adjust the resolution\\ of output images \end{tabular}
\\ 
\hline
IntuitiveIL & \begin{tabular}[l]{@{}l@{}} NeWCRFs \\ \citep{yuan2022new}\end{tabular} & T&SILog  & \begin{tabular}[l]{@{}l@{}}GC, DoG, \\normalization, HSV~\end{tabular} & \begin{tabular}[l]{@{}l@{}}AS, FA\end{tabular} &None              \\ 
\hline
\begin{tabular}[l]{@{}l@{}} KLIV  \end{tabular} & \begin{tabular}[l]{@{}l@{}} SUMNet \\ \citep{nandamuri2019sumnet} \end{tabular} & C&\begin{tabular}[c]{@{}c@{}}MAE + MSE\\ + SIL + BL \end{tabular} & Resizing, normalization  & & Low-pass GB         \\ 
\hline
\begin{tabular}[l]{@{}l@{}} MIVA  \end{tabular}       & \begin{tabular}[l]{@{}l@{}} DenseDepth \\ \citep{alhashim2018high}\end{tabular}  &C& MAE + SSIM &normalization &Horizontal flip &None \\ 
\hline
\begin{tabular}[l]{@{}l@{}} MMLab \end{tabular}       & \begin{tabular}[l]{@{}l@{}} Swin-UNet \\ \citep{cao2023swin} \end{tabular} &T &L1 &Downsampling &None &Upsampling
\\ 
\hline

\end{tabular}}\\
\scriptsize{T: Transformer backbone; C: Convolutional backbone; MAE: Mean Absolute Error; MSE: Mean Squared Error; BL: Berhu Loss; GC: Gamma Correction; DoG: Difference of Gaussian filter; AS: Average Shape; FA: Feature Augmentation; GB: Gaussian Blur; SSIM: Structural Similarity loss; SILog: Scale-Invariant Logarithmic loss; HSV: Hue Saturation Value.}

\label{table:methods_summary_subtask1}
\end{table*}

\begin{figure*}
    \centering
    \includegraphics[width=\linewidth]{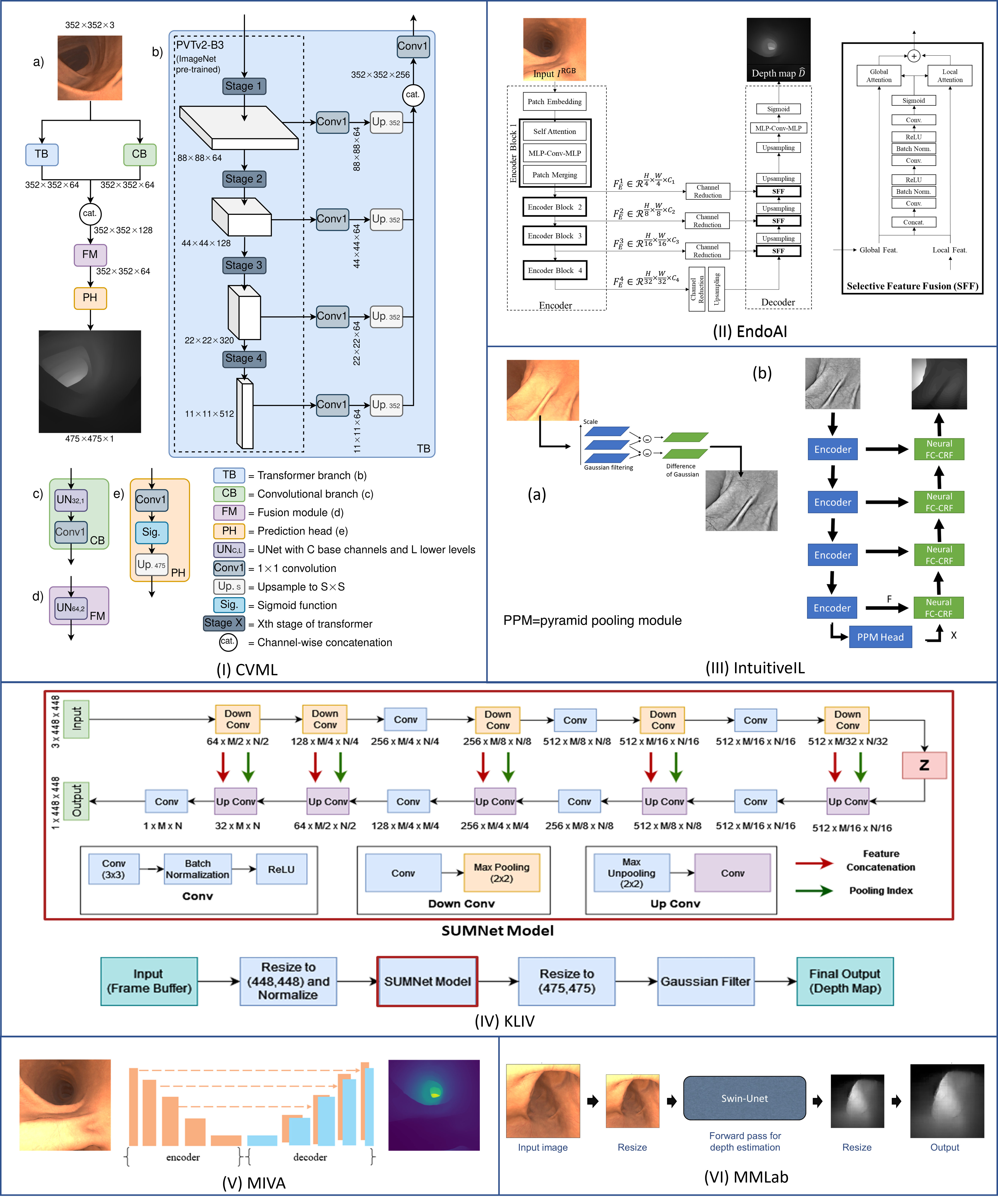}
    \caption{Architecture overview for Task 1 (depth prediction) of the 6 participating teams. (I) Team CVML adapted FCBFormer~\citep{sanderson2022fcn}, (II) Team EndoAI utilized GLPDepth~\citep{kim2022global} with Segformer encoder~\citep{xie2021segformer}, (III) Team IntuitiveIL applied multiple DoG filters with varying scales as preprocessing and used a NeW CRF network for depth prediction, (IV) Team KLIV utilized SUMNet \citep{nandamuri2019sumnet}, (V) Team MIVA utilized DenseDepth \citep{alhashim2018high} as an encoder-decoder network with skip connections, (VI) Team MMLab utilized Swin-UNet~\citep{cao2023swin}.}
    \label{fig:task1_block}
\end{figure*}
For Task 1, final submissions were received from six teams.
Table \ref{table:methods_summary_subtask1} summarizes the key features of the teams' methodology for Task 1. Team details and the methodology proposed by each participating team are presented below.

\subsection{FCBFormer adaptation by Team CVML}

Team CVML are Edward Sanderson and Bogdan J. Matuszewski from the University of Central Lancashire (UK). Team CVML proposed the FCBFormer-D (as shown in Fig.~\ref{fig:task1_block} (I)), which is an adaptation of the FCBFormer~\citep{sanderson2022fcn}. 

The overall architecture of FCBFormer-D is shown in Fig.~\ref{fig:task1_block}(I-a). The method consists of two branches: a transformer-based branch (TB) (Fig.~\ref{fig:task1_block}(I-b)) extracting global features, and a convolutional branch (CB) (Fig.~\ref{fig:task1_block}(I-c)) extracting local features that the TB could potentially neglect. For the Transformer branch, the Pyramid Vision Transformer v2 (PVTv2) \citep{wang2022pvt} (B3 variant pre-trained on ImageNet), which serves as image encoder and provides robust multiscale features for dense prediction, is employed. PVTv2 then feeds into a lightweight decoder. 
The convolutional branch is based on a UNet-style architecture inspired by \citep{nichol2021improved} and includes multi-head self-attention at the lower levels to provide the model with global context for this feature extraction. 
The feature maps from both branches are then concatenated and fused using a UNet-style architecture also inspired by \citep{nichol2021improved} in the fusion module (FM) Fig.~\ref{fig:task1_block}(I-d). 

Finally, the output of the fusion module is passed through the prediction head (PH) Fig.~\ref{fig:task1_block}(I-e). The prediction head is a 1x1 convolutional layer with sigmoid activation that outputs dense depth map. The depths are then upsampled to the original size of 475×475 using bilinear interpolation.

The implemented network takes a $352 \times 352$ RGB image with pixel intensities in the range $[-1, 1]$ as inputs. This involves resizing the $475\times 475$ 8-bit RGB images using bilinear interpolation with anti-aliasing prior to normalization. The output of the proposed network then provides a $475 \times 475$ depth map with relative depth values in the range $[0, 1]$. During training and validation, the ground truth depth values were scaled to a range of $[0, 1]$, corresponding to $[0$cm$, 20$cm$]$, and the model was optimized to minimize the mean squared error (MSE) loss. Team \textit{CVML} used AdamW optimizer with a learning rate of $1e-4$, which was scheduled to halve when the MSE on the validation data did not decrease over 10 epochs. The inputs were randomly horizontally and vertically flipped with a probability of 0.5. The model was trained for 300 epochs with a batch size of 24. The network weights with the smallest MSE on the validation set were saved. Training was performed on an ASUS ESC8000-G4 GPU server with six NVIDIA RTX A6000 48GB GPUs.

The groups' method is inspired by their observation that a standard UNet performs relatively weak at inferring the edges of the geometry, as well as the depth of far away surfaces. Their network thus aims to capture both, global features that help understand depth at all distances, and local features that can infer steps in depth.

\subsection{GLPDepth adaptation by Team EndoAI}
Team EndoAI are Jiwoon Jeon from EndoAI (Korea) and Jae Young Lee, Dong Jae Lee and Woonghyun Ka from Korea Advanced Institute of Science and Technology (Korea), who participated in all three Tasks. Team EndoAI proposed to use GLPDepth \citep{kim2022global}, a Transformer-based network for depth prediction (as shown in Fig.~\ref{fig:task1_block}(b), for the depth prediction task because this method has shown higher generalization ability and robustness compared to previously developed networks. To obtain the depth map prediction $D_{pred}$ from the input $I^{RGB}$, the local and global features are fused by Selective Feature Fusion (SFF) in the decoder. For the encoder, Segformer~\citep{xie2021segformer} is utilized.

The last layer of the original GLPDepth network decoder is modified to include a Conv2D-ReLU-Conv2D block to adjust the resolution of the resulting depth map. Further, to avoid scale adjusting, the model is directly trained to predict depth maps in the range of $[0, 1]$ (corresponding to $[0$cm$, 20$cm$]$) instead of using median scaling. GLPDepth uses the Scale-Invariant Logarithmic (SILog) loss \citep{eigen2014depth} given by:

\begin{equation}
    L(D_{pred}, D_{GT}) = \sqrt{\dfrac{1}{T}\sum_i d_i^2 -\left(\dfrac{1}{T}\sum_i d_i\right)^2}, 
\end{equation}
where $d_i$ is the pixel-wise log loss
\begin{equation}
    d_i = \text{log}(D_{pred}(i)) - \text{log}(D_{GT}(i))
\end{equation}
  and T denotes the number of pixels in the depth map.  
For training the GLPDepth model, the original hyperparameters from \citep{kim2022global} are used. The model is fine-tuned for 20 epochs using the \textit{CosineAnnealingWarmRestarts} learning rate scheduler \citep{loshchilov2016sgdr} on the challenge metrics: L1 depth error, RMSE, and relative depth error. The final model is chosen based on the performance of all metrics on the validation set. 

\subsection{NewCRFs adaptation by Team IntuitiveIL}

Team IntuitiveIL are Erez Posner, Netanel Frank, and Moshe Bouhnik from the Intuitive Surgical, who proposed to adapt Neural Window Fully-connected Conditional Random Fields (NeW CRFs) \citep{yuan2022new} to accomplish colonoscopy monocular depth estimation leveraging the advantages of fully-connected (FC) CRFs \citep{1315232}. In addition, they employed data augmentation techniques to address the issue of illumination changes, which involved creating partially illumination-invariant images.

For depth estimation, NeW CRFs are selected because they overcome the limitations of traditional depth estimation methods that rely on Markov Random Fields (MRFs) or CRFs~\citep{saxena2008make3d,NIPS2005_17d8da81}. NeW CRFs embed a vision transformer to capture pairwise interactions with multi-head attention as the encoder and the neural CRFs module in a network as the decoder. NeW CRFs can capture the relationship between any node in a graph, making them much stronger than neighbor CRFs. By splitting the input into windows and performing FC-CRFs optimization within each window, NeW CRFs reduce computation complexity while maintaining the advantages of FC-CRFs. Additionally, the use of multi-head attention within a neural CRFs module further improves depth estimation performance. As shown in Fig.~\ref{fig:task1_block}III(b), the encoder initially extracts features across four levels. A Pyramid Pooling Module (PPM) combines both global and local data, generating the preliminary prediction X using the uppermost image feature F. Subsequently, within each level, the neural window fully-connected CRF component constructs multi-head energy from X and F, refining it to an improved prediction X’. 

In colon augmentation, the method originally proposed in \citep{ye2014real} for face recognition is utilized, which contains the following steps to create the grayscale illumination-invariant image:
\begin{equation}\label{eq:colon aug}
    \begin{aligned}
    I_{gamma} &= GammaCorrection(\text{Image}) \\
    I_{DoG} &= DoG(I_{gamma}) \\
    I_{norm} &=\frac{I_{DoG}}{mean(|I_{DoG}|^a){\frac{1}{a}}}\\
    I_{norm} &= \frac{I_{norm}}{mean(min((\tau),|I_{norm}|^a))^\frac{1}{a}}\\
    I_{norm} &= \frac{\tau *tanh(I_{norm})}{\tau},
     \end{aligned}
\end{equation}
where $GammaCorrection$ involves gamma correcting all images to the same value, and $DoG$ represents the difference of Gaussians filter. In the augmentation process, the original DoG image is replaced with an average of several DoG filters with varied scales (as illustrated in Fig.~\ref{fig:task1_block}III(a)). This augmentation aims to improve local texture and accommodates features of various sizes. Additionally, the input image is changed from an RGB to an HSV representation, the value channel is swapped for the algorithm's output in grayscale, and the resulting image is then converted back to an RGB representation. This allowed stronger features even in the colon's distant areas. Scale-Invariant Logarithmic (SILog) loss is utilized as the loss function. SILog supervises the training by first calculating the logarithm difference between the predicted and the ground-truth depth map. For K pixels with valid depth values in an image, the scale-invariant loss is computed to measure the performance of the depth estimation \citep{yuan2022new}. 
\subsection{SUMNet adaptation by Team KLIV}
Team KLIV are Varshini Elangovan from College of Engineering, Guindy (India), and Sista Raviteja, Rachana Sathish, Debdoot Sheet from the Indian Institute of Technology Kharagpur (India). 
KLIV proposed to apply a fully convolutional neural network SUMNet \citep{nandamuri2019sumnet} to effectively generate colon depth maps from frame buffers while preserving conformity around small structures and preventing the loss of critical information. 

Concretely, SUMNet \citep{nandamuri2019sumnet} consists of an encoder network with VGG11 architecture, activation concatenation, and pooling index transfer. Several loss functions are taken into account during the training process, including Mean Absolute Error (MAE), Mean Squared Error (MSE), scale-invariant loss \citep{eigen2014depth}, and Berhu loss \citep{carvalho2018regression}. 
In order to reduce the aliasing effect in the predicted depth maps, a post-processing step is used to apply a Gaussian Blur low-pass filter with a kernel size of $7 \times 7$. 

From the provided training data 10,309 frames are used for training, and 3,603 frames are used for validation. To give a more thorough summary, the frame buffers in the simulated dataset were initially in RGBA format, but for network compatibility, they are converted to RGB images and resized to 448 × 448. Additionally, the images are normalized using the training dataset's mean and standard deviation. The depth maps are scaled to 448 × 448 and translated into grayscale images. These preprocessed images and depth maps are then used for training the SUMNet model for depth estimation of synthetic colonoscopic images. The network is implemented in PyTorch and trained for 50 epochs on an Nvidia GeForce GTX TITAN X GPU with a batch size of 16 using the ADAM optimizer with an initial learning rate of 0.001 and an exponential learning rate scheduler with a decay factor of 0.98. The complete training took 24 hours. 

The effectiveness of the model and the reliability of its predictions are assessed using the L1 error, relative error, and root-mean-square error. The model trained on MSE loss predicted results that are more reliable and accurate, in comparison to the models trained on the other loss functions. KLIV's code is available\footnote{\url{https://github.com/SistaRaviteja/Colonoscopy-Depth-Estimation}}.

\subsection{DenseDepth adaptation by Team MIVA}
Team MIVA are Zhengwen Li and Yichen Zhu from ZheJiang University (China), who participated in all three Tasks. MIVA used DenseDepth \citep{alhashim2018high} which is a fully convolutional encoder-decoder architecture with skip connections (as shown in Fig.~\ref{fig:task1_block}(V)). The encoder is a DenseNet-169~\citep{huang2017densely} pre-trained on ImageNet \citep{deng2009imagenet} as proposed by the original DenseDepth. The authors also experimented with a DenseNet-201, which performed worse in their experiments.
To train the network, MIVA used the loss $L$ as the weighted sum between the depth and SSIM loss:
\begin{equation}
    L(Y,Y')=0.1\cdot L_{depth}(Y,Y')+L_{SSIM}(Y,Y').
\end{equation}
The Loss term $L_{depth}$ is the point-wise L1 loss defined on the depth values and $L_{SSIM}$ uses the Structural Similarity (SSIM). The authors replace the original augmentation strategy with a 50\% random horizontal flipping and image normalization only. The synthetic data provided is split into training set\citep{rau2022bimodal} and validation set in the way recommended by the SimCol3D challenge organizers, and the mean and standard deviation in normalization are calculated from all images in the training set. The participants trained their method on an NVIDIA GeForce RTX 3090 GPU using a batch size of 16 and a learning rate of $10^{-4}$ with Adam optimizer for 40 epochs. 

\subsection{Swin-UNet adaptation by MMLAB}
Team MMLAB are Seenivasan Lalithkumar, Islam Mobarakol and RenHongliang are from National University of Singapore (Singapore), Imperial College London (UK) and Chinese University of Hong King (China), who participated in Task 1 and 2.  

For the depth estimation task, a Unet-like Swin-Transformer (Swin-UNet)~\citep{cao2023swin} (Fig.~\ref{fig:task1_block}(VI)), a medical image segmentation model, is used. Swin-UNet forms a hierarchical Swin Transformer with shifted windows in the encoder, a decoder with patch expanding layer to perform upsampling on the feature maps and skip connections for local-global semantic feature learning. Overall, there are three blocks of the encoder and corresponding decoder in Swin-UNet. The model was trained using L1 loss and SGD optimizer with a learning rate of 0.01, a decay factor of 1e-4, and a momentum of 0.9. The input images are resized to $224 \times 224$ during training and upsampled to the original size at test time after the prediction. The participants experimented with different loss functions such as L1, mean square error (MSE), structural similarity index (SSIM), and binary cross entropy (BCE). Ultimately, the L1 loss outperformed other loss functions with an MSE of 0.000115 and an SSIM of 0.984670. The team's code is publicly available\footnote{\url{https://github.com/lalithjets/SimCol3D_challenge_2022}}.

\section{Methods for Task 2 and 3: Pose prediction from synthetic and real images}
\begin{figure*}[ht!]
    \centering
    \includegraphics[width=1\linewidth]{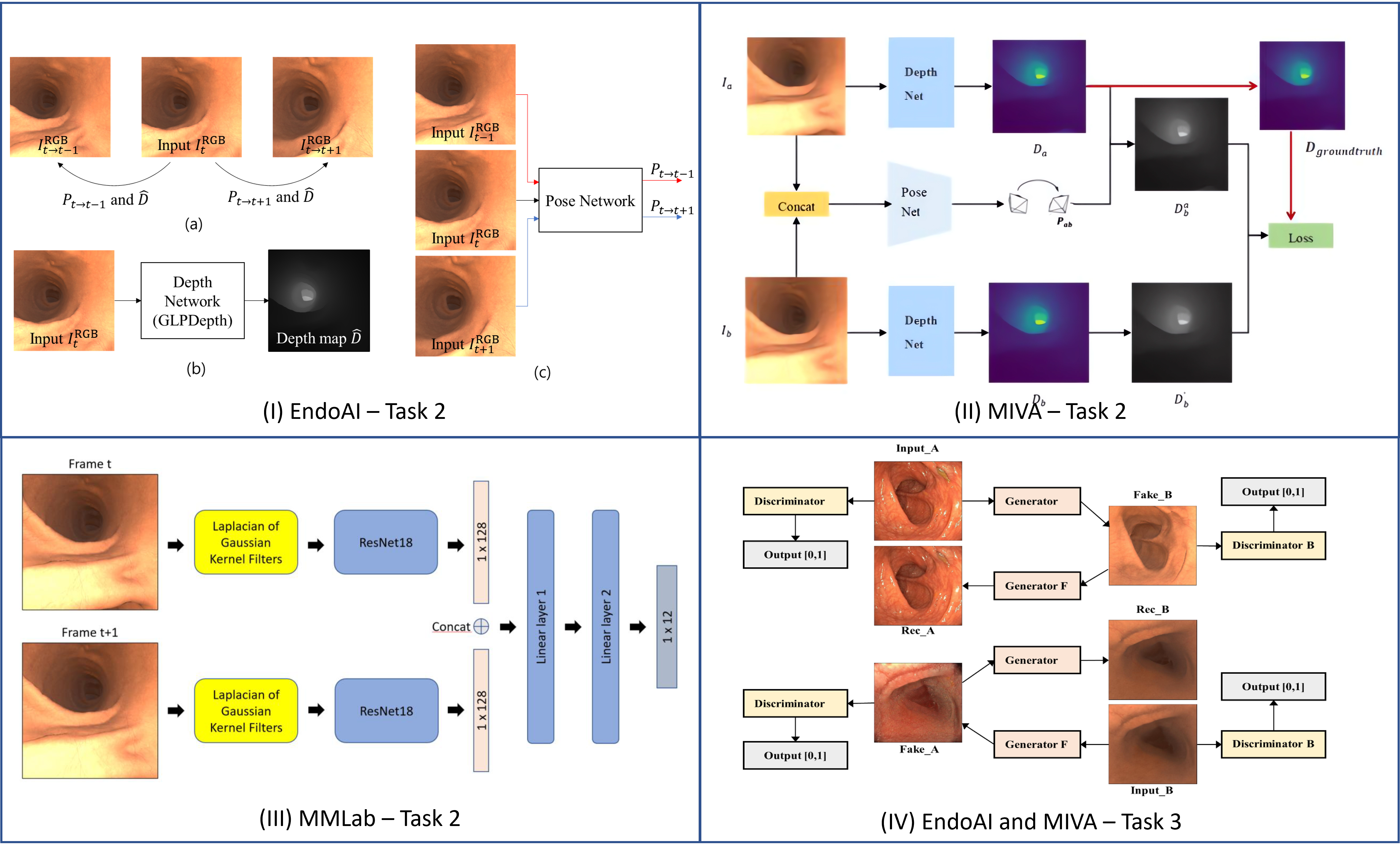}
    \caption{Architecture overview for Task 2 (pose prediction from synthetic) and Task 3 (pose prediction from real) images of the participating teams. For Task 2, (I) Team EndoAI utilized MonoDepthv2~\citep{godard2019digging}, (II) Team MIVA utilized SC-SfMLearner~\citep{bian2021unsupervised}, and (III) Team MMLab implemented curriculum learning with linear regression. For Task 3, (IV) Team EndoAI and Team MIVA utilized the CycleGAN model for Sim2Real image generation.}
    \label{fig:task2_3_block}
\end{figure*}

\begin{table}
\caption{Summary of the participating teams of the SimCol3D Challenge – Task 2 and Task 3}
\centering
\resizebox{\columnwidth}{!}{
\begin{tabular}{l|l|l|l|l} 
\hline
\multirow{1}{*}{\textbf{Team}} & \multicolumn{1}{c|}{\textbf{Task}} & \multicolumn{1}{c|}{\textbf{Algorithm}} & \multicolumn{1}{c|}{\textbf{Loss function}}    & \multicolumn{1}{c}{\textbf{Data augmentation}}            \\
\hline
\multirow{2}{*}{EndoAI} &2 (Pose Syn.) &\begin{tabular}[l]{@{}l@{}}MonoDepth2 \\\citep{godard2019digging} \end{tabular}  &MSE &None \\ 
\cline{2-5} 
& 3 (Pose Real)  &\begin{tabular}[l]{@{}l@{}}CycleGAN + \\MonoDepth2 \end{tabular} &Same as Task 2 &\begin{tabular}[l]{@{}l@{}} CycleGan real-to-syn \\ conversion \end{tabular} \\ 
\hline
\multirow{2}{*}{MIVA} &2 (Pose Syn.) &\begin{tabular}[l]{@{}l@{}}SC-sfMLearner \\ \citep{bian2021unsupervised} \end{tabular} &\begin{tabular}[l]{@{}l@{}}SC-sfMLearner  \\+ Densedepth \end{tabular} &Image normalization \\ 
\cline{2-5} 
& 3 (Pose Real)  &\begin{tabular}[l]{@{}l@{}}CycleGAN + \\SC-sfMLearner \end{tabular} &Same as Task 2 &\begin{tabular}[l]{@{}l@{}}Crop, resize, CycleGAN  \\ real-to-syn conversion \end{tabular} \\ 
\hline 
\multirow{2}{*}{MMLab} &2 (Pose Syn.) &\begin{tabular}[l]{@{}l@{}}Curriculum learning,\\ linear regression \end{tabular} &MSE &None\\ 
\cline{2-5} 
& 3 (Pose Real)  &N/A  &N/A  &N/A  \\ 
\hline

\end{tabular}

}
\label{table:methods_summary_task2-3}
\end{table}

In total, 3 teams (EndoAI, MIVA and MMLab) participated in Task 2 (pose prediction from synthetic), two of which (EndoAI and MIVA) also participated in Task 3 (pose prediction from real images). Table \ref{table:methods_summary_task2-3} provides an overview of the key features of the teams' methodology. The remainder of this section describes the participants' methods in detail.

\subsection{SC-SfMLearner adaptation by MIVA}
For the pose estimation task, MIVA used a method based on SC-SfMLearner as shown in Fig.~\ref{fig:task2_3_block}(II), which includes two parts: a depth estimation module and a pose estimation module. In addition, they replaced the DispResNet depth estimation module in the original SC-SfMLearner with a DenseDepth network. As ground truth depth for synthetic data was known, MIVA made use of this information while training the formerly self-supervised SC-SfMLearner. To supervise the depth module, the loss of Densedepth was added to the original loss of SC-SfMLearner. The modified loss function is 
\begin{equation}
    L=L_{SC-sfmformer} + \omega\cdot L_{densedepth},
\end{equation}
where the weight $\omega$ was set to 1.

The team divided the dataset according to their split for Task 1 and also normalized the input images. MIVA's model was trained on an NVIDIA GeForce RTX 3090 GPU with a batch size of 8, learning rate of $10^{-4}$ and Adam optimizer. The network was trained for 40 epochs.

For the Task 3,  MIVA used CycleGAN, which consists of two generators and two discriminators as shown in Fig.~\ref{fig:task2_3_block}(IV), where $A$ represents the real colonoscopy image domain, and $B$ represents the virtual colonoscopy image domain. The input image A generates Fake\_B through Generator G, and Fake\_B generates Rec\_A through Generator F. After two transformations, Rec\_A is mapped back to the A domain. The model is optimized by comparing the similarity between Input\_A and Rec\_A. Input\_B is processed in the same way. The generator in this paper adopts a ResNet backbone~\citep{he2016deep}, and the discriminator uses a PatchGAN structure. The EndoMapper~\citep{azagra2023endomapper} dataset and the synthetic dataset~\citep{rau2022bimodal} provided by the SimCol3D Challenge were used for training the CycleGAN. Since there are black areas in the four corners of the EndoMapper dataset, MIVA cropped the areas from (155, 0) to (1162, 1007) and reduced them to a $480\times480$ square. During training, MIVA applied random horizontal flipping and normalization to the data. Preliminary validation results show that CycleGAN's generator can map multiple inputs to the same output. For example, a real colonoscopy image is converted to generate a completely different virtual image. For this reason, MIVA experimented with identity loss, self-regularization loss and SSIM loss to guide the generator. Ultimately, the team used an identity loss in the final submitted model. The experiments for Task 3 was carried out on an NVIDIA GeForce RTX 2080ti.

\subsection{MonoDepth2 adaptation by EndoAI}
\textit{EndoAI's} camera pose estimation framework is based on the self-supervised monocular depth estimation method called MonoDepth2 \citep{godard2019digging} (as shown in Fig.~\ref{fig:task2_3_block}(I-a)) but is trained using supervision with the ground truth translations and rotations. In addition to the original self-supervised loss, the team added the supervised loss
\begin{equation}
    L(P_{pred}, P_{GT})=\sum_{i,j} ||P_{pred} - P_{GT}||_1,
\end{equation}
where $P_{pred}$ and $P_{GT}$ are the prediction and ground truth $4 \times 4$ matrices, and $(i, j)$ represents row and column indices of the matrices, respectively, such that $(1 \leq i \leq 3,1 \leq j \leq 4)$. The last row is not used. 

Further, the depth network in Monodepth2 is replaced with GLPDepth \citep{kim2022global} (Fig.~\ref{fig:task2_3_block}(I-b)) and the pose network employs a ResNet18 encoder and a decoder (Fig.~\ref{fig:task2_3_block}(I-c)). For Task 2, the depth network is trained from scratch (weights from Task 1 are not used). The team used the hyperparameters proposed in the original MonoDepth2. At training time, as used in Monodepth2, both forward and backward path trajectories are trained, simultaneously. The model is trained for 20 epochs, but the epoch with the best performance on the validation set is submitted. The same model is used for Task 3, where additionally, a CycleGAN was used to translate real images to synthetic images before feeding into the Monodepth2 network. 

\subsection{Curriculum learning with linear regression by MMLab}
For Task 2, MMlab employed ResNet18~\citep{he2016deep} and a series of linear layers as shown in Fig.~\ref{fig:task2_3_block}(III). Furthermore, they employed Laplacian of Gaussian (LoG) kernel-based filters to enforce attention to contours and perform curriculum learning. Initially, the ResNet module is loaded with the PyTorch ImageNet pre-trained weights. Then the whole model is trained based on mean-square-error (MSE) loss using Adam optimizer with a learning rate of $7.5 \times 10^{-6}$ for 45 epochs. During training, the values of the LOG kernel (with kernel size = 3) are updated with a factor of 0.9 to allow more features to pass through the model as the learning progresses and to enforce attention to contours. While the relative ground truth pose has 16 values, the module regresses 12 values as the last four values are constant [0.0, 0.0, 0.0, 1.0].

\section{Results and discussions}
This section summarizes and discusses the submitted results of all participating teams on the three tasks. 
\subsection{Task 1: Depth estimation}

\begin{table*}
    \centering
        \caption{Task 1 results on the three test scenes. We report the mean over three sequences per test scene. Winners are indicated in bold, the runner-up is underlined, and third-placed teams are shown in italics. Asterisks (*) indicate scenes that provided trajectories with groundtruth for training. All results are reported in cm.}
    \begin{tabular}{l|c|c|c|c|c|c|c|c|c|c}
    \hline
         & \multicolumn{3}{c|}{SynCol I*} & \multicolumn{3}{c|}{SynCol II*} & \multicolumn{3}{c|}{SynCol III}& \multirow{2}{*}{$\sum_1\uparrow$}\\
        \cline{2-10}
         &L1 $\downarrow$ & Rel $\downarrow$& RMSE $\downarrow$& L1 $\downarrow$ & Rel $\downarrow$& RMSE $\downarrow$& L1 $\downarrow$& Rel $\downarrow$ & RMSE $\downarrow$& \\
         \hline \hline
CVML&\textbf{0.030}&\textbf{0.012}&\textbf{0.045}&\textbf{0.030}&\textbf{0.009}&\textbf{0.044}&\textbf{0.099}&\textbf{0.025}&\textbf{0.141}& \textbf{54}\\
EndoAI&\textit{0.040}&\textit{0.015}&\textit{0.067}&\textit{0.039}&\textit{0.011}&\underline{0.063}&\textit{0.111}&\textit{0.028}&\textit{0.168}&\textit{37}\\
IntuitiveIL&0.050&0.017&0.091&0.059&0.016&0.103&0.167&0.047&0.233&26\\

KLIV&0.155&0.055&0.228&0.166&0.045&0.236&0.187&0.048&0.277&12\\
MIVA&\underline{0.038}&\underline{0.014}&\underline{0.065}&\underline{0.038}&\underline{0.010}&\textit{0.065}&\underline{0.107}&\underline{0.025}&\underline{0.163}&\underline{44}\\
MMLAB&0.109&0.037&0.185&0.201&0.047&0.330&0.171&0.040&0.277&16\\
         \hline    
    \end{tabular}
    \label{tab:res_task1}
\end{table*} 

\begin{figure*}
    \centering
    \includegraphics[width=0.9\linewidth]{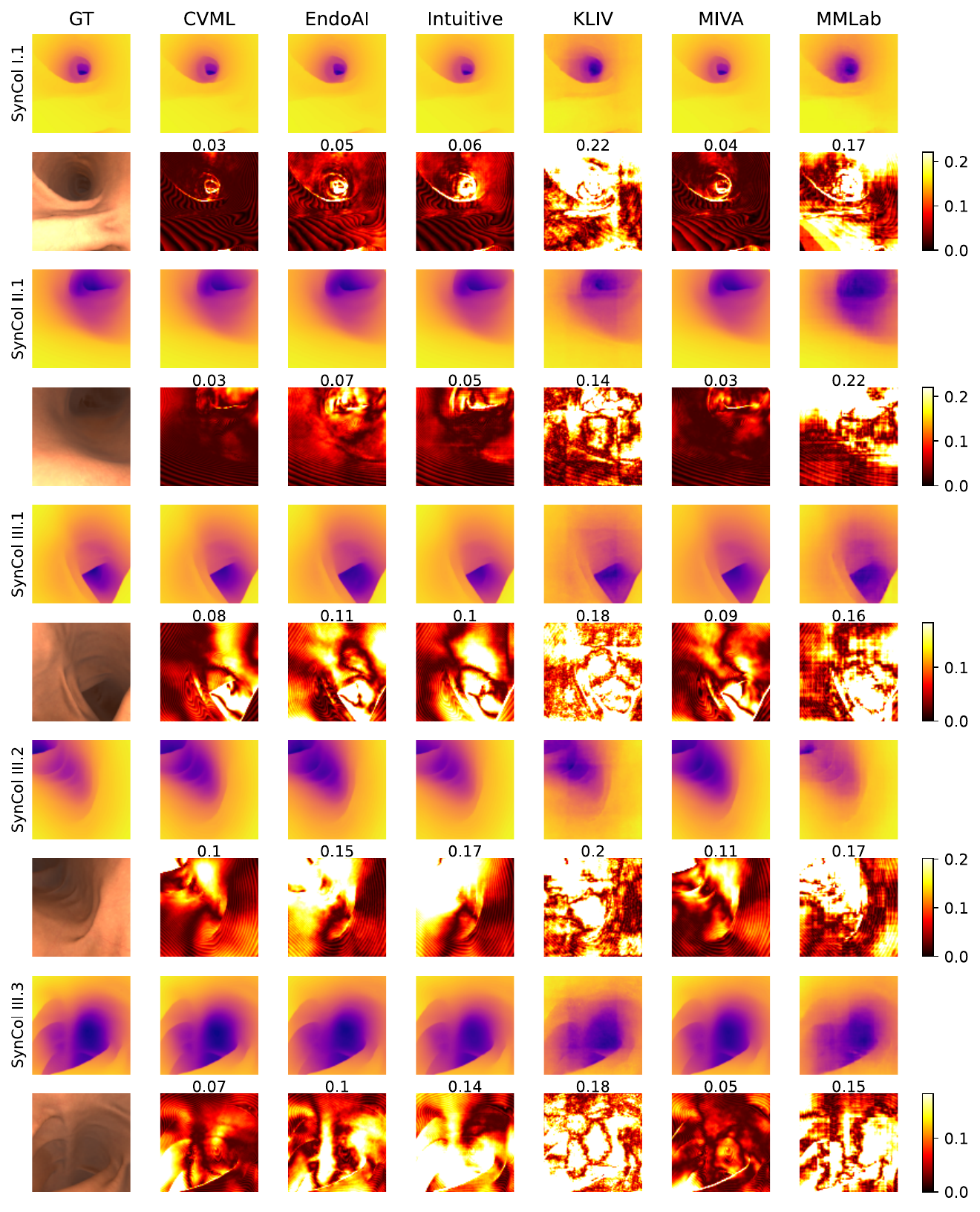}
    \caption{ Comparison of depth predictions generated by the participant teams. For Synthetic Colons I and II we show one example from one test trajectory each. For Synthetic Colon III, we show an example for all three test trajectories. We show the average L1 error above each error map. The colorbar's scale is in cm. Visually, the results of CVML, EndoAI, IntuitiveIL, and MIVA are barely distinguishable from the ground truth. Though when observing the L1 error, CVML is found to be the best performing one, closely followed by MIVA.  }
    \label{fig:depth_comp}
\end{figure*}
\begin{table*}
    \centering
        \caption{Task 2 results on the three test scenes. We report the mean over three sequences per test scene. Winners are indicated in bold, and the runner-up is underlined. ATE is measured in dm, RTE in cm, and ROT in degrees. Asterisks (*) indicate scenes that provided trajectories with ground truth for training. }
    \begin{tabular}{l|c|c|c|c|c|c|c|c|c|c}
    \hline
         & \multicolumn{3}{c|}{SynCol I*} & \multicolumn{3}{c|}{SynCol II*} & \multicolumn{3}{c|}{SynCol III}&\multirow{2}{*}{$\sum_2$}\\
         \cline{2-10}
         &ATE  $\downarrow$ & RTE  $\downarrow$& ROT  $\downarrow$& ATE  $\downarrow$ & RTE  $\downarrow$& ROT  $\downarrow$ & ATE  $\downarrow$ & RTE  $\downarrow$& ROT $\downarrow$ \\
         \hline \hline
EndoAI&\textbf{0.574}&\textbf{0.081}&\underline{0.144}&\underline{0.336}&\textbf{0.084}&\textbf{0.148}&\textbf{0.325}&\underline{0.247}&\underline{0.367}&\textbf{0.165}\\
MIVA&0.860&0.124&\textbf{0.141}&\textbf{0.325}&0.158&\underline{0.180}&\underline{0.422}&\textbf{0.226}&\textbf{0.275}&\underline{0.183}\\
MMLAB&\underline{0.819}&\underline{0.082}&2.818&1.206&\underline{0.139}&1.880&0.572&0.458&1.833&0.284\\

         \hline
         
    \end{tabular}
    \label{tab:res_task2}
\end{table*}

\begin{figure*}
    \centering
    \begin{tabular}{lccc}
    & EndoAI & MMLAB & MIVA \\

     \rotatebox[origin=c]{90}{SynCol I}&\raisebox{-0.5\height}{\includegraphics[width=0.25\textwidth]{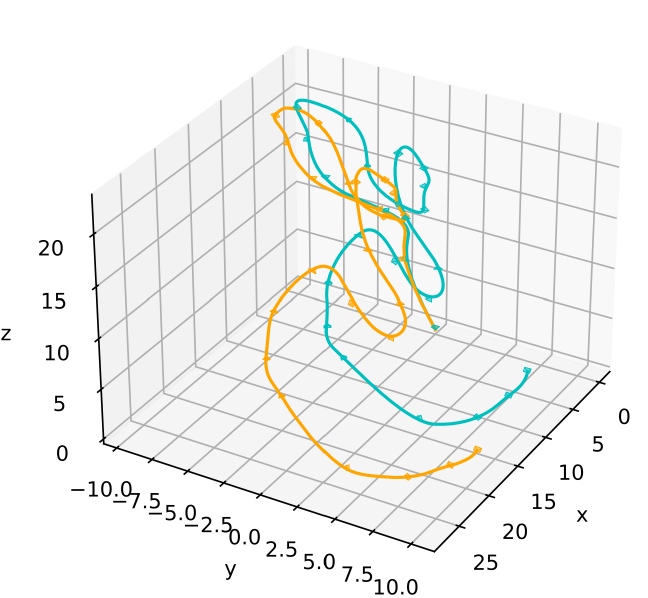}}   & \raisebox{-0.5\height}{\includegraphics[width=0.25\textwidth]{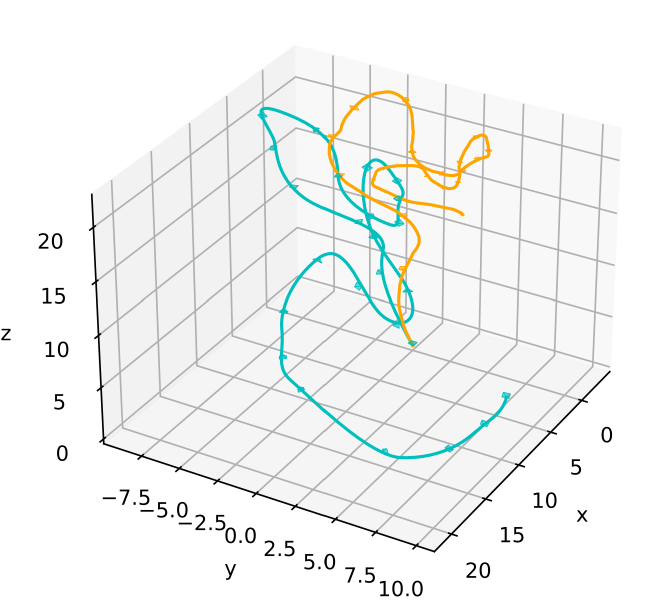}} &\raisebox{-0.5\height}{\includegraphics[width=0.25\textwidth]{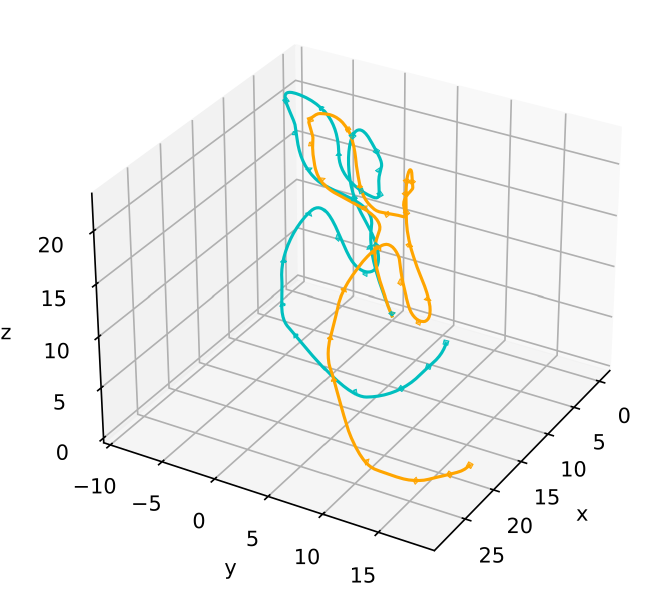}}\\
      \rotatebox[origin=c]{90}{SynCol II}&\raisebox{-0.5\height}{\includegraphics[width=0.25\textwidth]{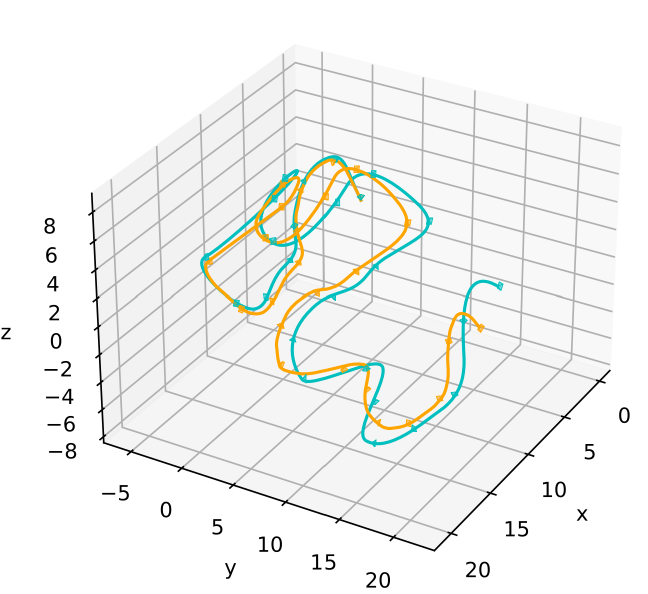}}   & \raisebox{-0.5\height}{\includegraphics[width=0.25\textwidth]{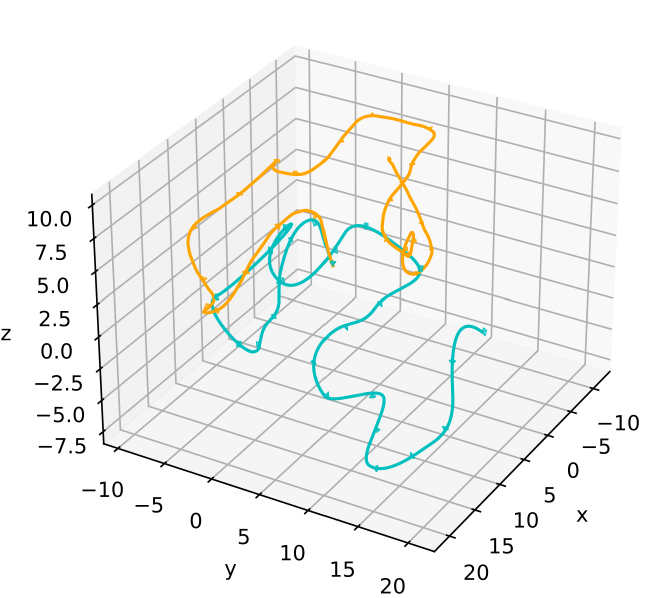}} &\raisebox{-0.5\height}{\includegraphics[width=0.25\textwidth]{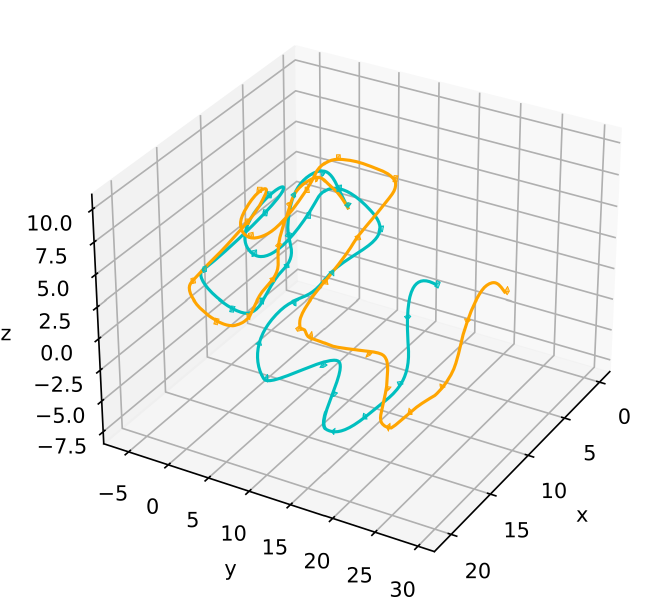}}\\
      \rotatebox[origin=c]{90}{SynCol III}&\raisebox{-0.5\height}{\includegraphics[width=0.25\textwidth]{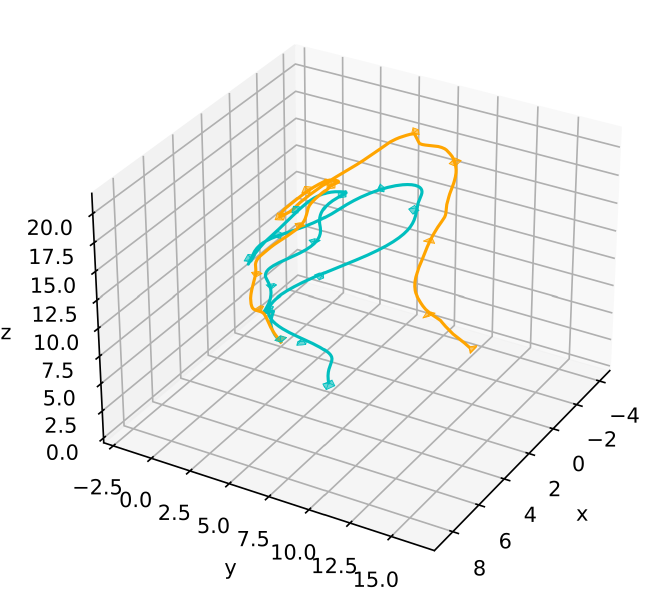}}   & \raisebox{-0.5\height}{\includegraphics[width=0.25\textwidth]{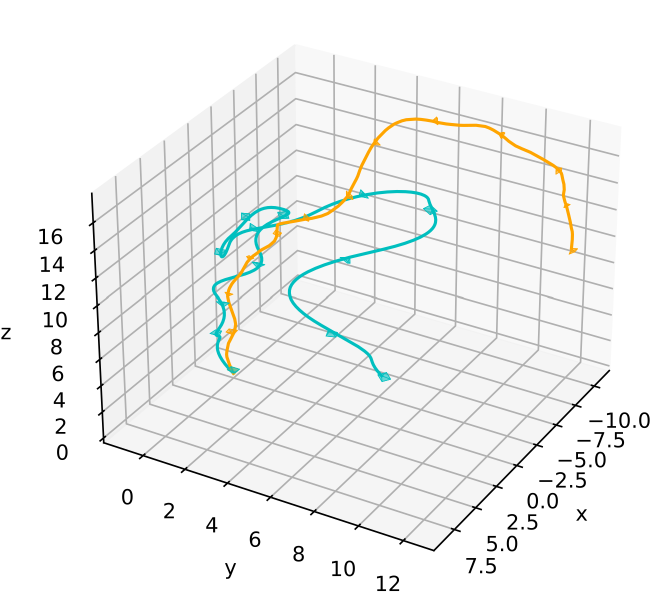}} &\raisebox{-0.5\height}{\includegraphics[width=0.25\textwidth]{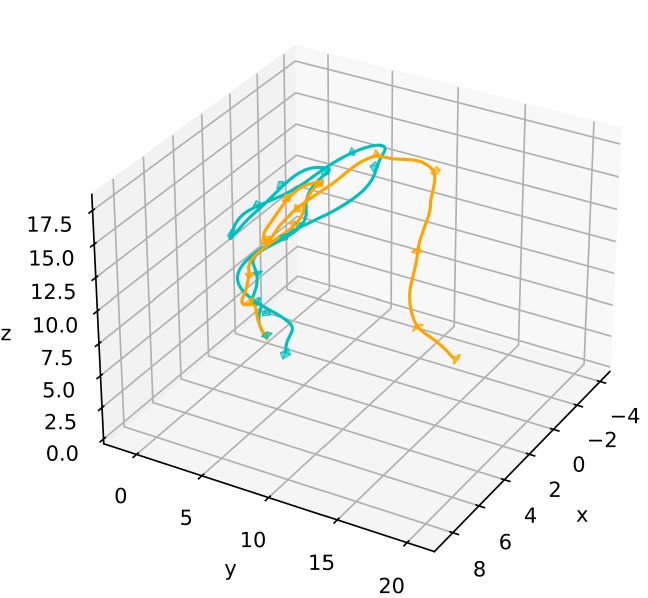}}\\

    \end{tabular}
    \caption{Overview of Task 2 results showing \color{orange} predicted \color{black}    and \color{cyan} ground truth \color{black}   trajectories. For each test scene, we show the first of three trajectories. Qualitatively, it can be observed that Team EndoAI performed the best.}
    \label{tab:traj_t2}
\end{figure*}
All teams that participated in Task 1 delivered impressive results on the test scenes as presented in Table \ref{tab:res_task1}. The $L_1$ error ranged between $0.03$ cm and $0.201$ cm across teams and scenes. 

Among the three best-performing methods, one method was fully convolutional (MIVA), and the other two were a combination of a convolutional model and transformer (EndoAI, CVML). 
Achieving sub-millimeter errors on all scenes, team CVML demonstrates that depth prediction from synthetic data can be considered a robustly solvable task. CVML outperformed all other teams on all metrics and all scenes. Even on SynCol III, a scene that has not been seen during training, the average $L_1$ error of CVML is below one millimeter. The winning team used a model that combines both a transformer-based and a CNN-based branch in a single network. To develop their method, the team performed detailed validation of their backbone model, which inspired their modifications to FCBFormer. The team reported that the addition of the $1\times1$ convolutional layers to the convolutional branch and the inclusion of the fusion module was instrumental to their method's accuracy. Furthermore, the multi-head self-attention in both the U-Net style architecture in the convolutional branch and the fusion module boosted performance but necessitated replacing the Transformer branch decoder and the prediction head with lightweight alternatives to reduce computational complexity. 
 
Maybe surprisingly, the runner-up method proposed by MIVA is based on a convolutional neural network from 2018 and was applied out of the box, without further adaptations to the method, or complex augmentations or post-processing. The only changes the authors made was replacing the original augmentations with less pronounced endoscopy-suitable augmentations, namely flipping and normalizing only, which might be a good strategy for synthetic colonoscopy frames as their appearance does not vary. 

  EndoAI's method ranked third and was also a direct adaptation of an existing method, but, like the winner, the model is based on a combination of a convolutional model and a Transformer.

Comparing all six methods, we find that all teams used distinctly different baseline methods. Perhaps surprisingly, there was no consensus on the best method for depth prediction during the time of the challenge. Most teams used recent works from 2022 to build upon except for two teams, one of which ranked second. All recent methods were transformer-based, while the older methods are CNNs. Team IntuitiveIL was the only team to develop new augmentation strategies tailored towards colonoscopy applications. The team introducing most changes to a baseline method is CVML which won the first task of the challenge.

  Interestingly, the winning method and the last and second-to-last methods employed networks that were initially developed for medical image segmentation. All other teams used networks that were developed for depth prediction. Given the discrepancy between the results, it appears that in this challenge, segmentation models are neither better nor worse than depth prediction networks. 

Similarly, fully convolutional networks ranked both second, and last, so that a method's performance cannot be attributed to this design choice alone. Transformer-only networks ranked fourth and fifth, suggesting, that perhaps, the transformers used in this challenge were not equipped to capture the detailed geometry of the endoscopic scenes. Although Vision Transformers have greatly impacted the broad field of computer vision, further investigations into their ability to predict depth from endoscopic images are required. One design, that performed well throughout, is a combination of transformer and convolutional layers. As described by Sanderson et al. \citep{sanderson2022fcn}, who also participated in this challenge as team CVML, the combination of transformer and convolutional layers helps leverage both global and local features in endoscopic images. 

Comparing quantitative results in Table \ref{tab:res_task1}, we can observe that three best performing methods all lead to similar errors. For instance, on SynCol I, CVML, EndoAI, and MIVA achieve L1 errors of $0.03 - 0.04$ cm. The other methods perform considerably worse ($0.05 - 0.16)$. Further, all methods perform significantly worse on SynCol III, for which, as opposed to SynCols I and II, there were no training sequences released. Nonetheless, all methods achieve an L1 loss of less than 2mm. This speaks to the ability of these methods to accurately generalize to unseen geometries.   

A qualitative comparison of all methods on a few representative images from all three scenes is provided in Figure \ref{fig:depth_comp} along with the L1 error of individual predicted masks. We randomly sampled one frame for visualization per trajectory. The results of CVML, EndoAI, IntuitiveIL, and MIVA are barely distinguishable from the ground truth. Only when assessing the individual L1 errors, we can observe that CVML performs slightly better than MIVA, followed by EndoAI and IntuitiveIL. KLIV and MMLab show visible checkerboard artefacts, which is consistent with the quantitative results in Table \ref{tab:res_task1}, where KLIV and MMLab rank fifth and sixth.

\subsection{Task 2: Camera pose estimation on simulated data}

Three of the six teams participated in Task 2. The results of these teams are summarized in Table \ref{tab:res_task2}. The challenge organizers were particularly interested in the teams' results on the third test scene, as no training trajectories of scene III were provided to the teams. We thus weighted errors on SynCol III twice, while errors on SynCol I and II were weighted once, to reflect the importance of generalizability to unseen scenes. Based on the mean ATE, Team EndoAI performs best and by a large margin on two out of three trajectories and takes first place. EndoAI also performs best on SynCol III, which is the only unseen scene. Based on the RTE, EndoAI performs best on SynCol I and II, but even when weighting results on scene III twice, EndoAI outperforms the other methods. MIVA performs best on all measures in at least one scene, but ranks second overall, followed by MMLAB.

Qualitative results are shown in Figure \ref{tab:traj_t2}. We chose to show one trajectory per scene only, as the differences between trajectories on one scene are small. 
It can be observed that EndoAI's predictions most closely follow the ground truth trajectories. All models show clear drift in almost all scenes, which is consistent with the frame-wise approaches all teams chose to follow. Especially scene SynCol III, which was not seen during training, suffers from drift. Notably, the two more accurate approaches are both based on warping-based depth and pose networks (MonoDepth2 with updated backbones and SC-SfMLearner), while the third-placed method regresses pose from images directly. Although the warping-based approaches are optimized for the auxiliary task novel-view synthesis, the networks outperform the approach that minimized the pose loss only. Moreover, the two teams employing warping-based networks added supervised losses based on the provided labels in the training to the respective self-supervision methods. Interestingly, MIVA employed a supervised depth loss, while EndoAI used a supervised pose loss, and neither team used both depth and pose labels. As all teams use different backbones, a concluding comparison study remains to be conducted. We can only speculate that EndoAI's performance might result from their more complex back-bone (Transformer-based depth network) in comparison to MIVA who use a UNet-type depth net. It could also result from their supervision with ground truth poses in addition to the self-supervised losses of Monodepth2. 
\begin{table} 
    \centering
        \caption{Task 3 results. Winners are indicated in bold. The ROT error is reported in degrees. The absolute scale of the ATE and RTE is unknown.}
    \resizebox{1.0\columnwidth}{!}{
    \begin{tabular}{l|c|c|c|c|c|c|c|c}
         \hline
    Sequence     & 1&2&3&4&5&6&7&\multirow{2}{*}{$\sum_3$}\\
         \cline{1-8}
 \#Frames/seq & 76 &144&119&69&127&86&56&\\
          \hline
         \hline
& \multicolumn{7}{c|}{ATE  $\downarrow$}&ATE  $\downarrow$\\

\hline
EndoAI & 3.34&10.19&7.70&\textbf{1.17}&\textbf{1.47}&11.58&14.69& 7.16\\
MIVA& \textbf{0.97}&\textbf{4.50}&\textbf{3.12}&2.38&{3.55}&\textbf{4.10}&\textbf{6.48}&\textbf{3.59}\\
         \hline
                  \hline
& \multicolumn{7}{c|}{RTE  $\downarrow$}&RTE  $\downarrow$\\
\hline
EndoAI & {0.104}&{0.200}&{0.142}&{0.307}&{0.144}&\textbf{0.191}&\textbf{0.493}&0.23\\
MIVA& \textbf{0.065}&\textbf{0.104}&\textbf{0.130}&\textbf{0.174}&\textbf{0.116}&0.300&0.625&\textbf{0.22}\\
         \hline
                  \hline
& \multicolumn{7}{c|}{ROT  $\downarrow$}&ROT  $\downarrow$\\
\hline
EndoAI & {0.709}&{0.960}&{0.836}&{0.643}&{0.776}&{0.823}&{1.310}&0.87\\
MIVA& \textbf{0.264}&\textbf{0.634}&\textbf{0.551}&\textbf{0.453}&\textbf{0.622}&\textbf{0.478}&\textbf{0.804}&\textbf{0.54}\\
         \hline

         \hline        
    \end{tabular}
}
    \label{tab:res_task3}
\end{table}

\begin{figure*} 
    \centering
    \setlength\tabcolsep{1pt}
    \begin{tabular}{ccc}
    
    \vspace{2em}
    
    \begin{minipage}{0.3\linewidth} \centering 1\\ \includegraphics[width=\textwidth]{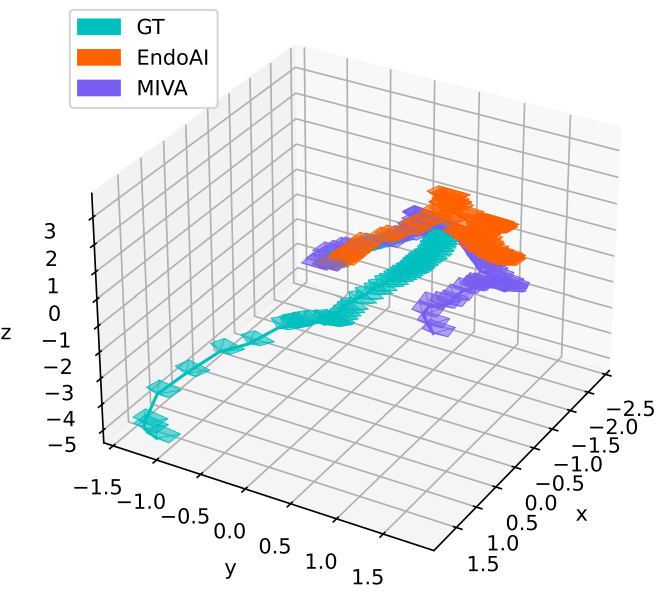} \\ \includegraphics[width=0.25\textwidth]{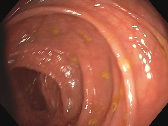} \includegraphics[width=0.25\textwidth]{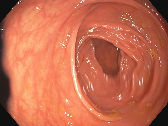} \includegraphics[width=0.25\textwidth]{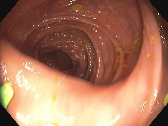}\end{minipage}
    
    &\begin{minipage}{0.3\linewidth} \centering 2 \includegraphics[width=\textwidth]{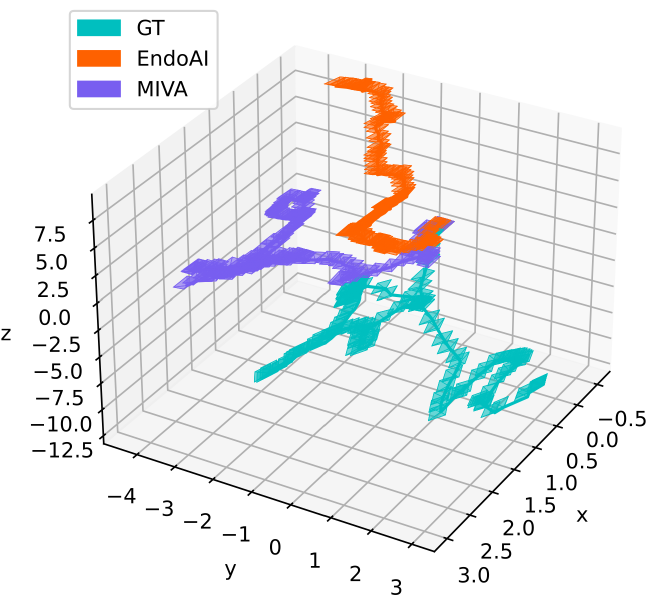} \\ \includegraphics[width=0.25\textwidth]{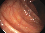} \includegraphics[width=0.25\textwidth]{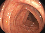} \includegraphics[width=0.25\textwidth]{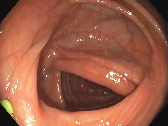}\end{minipage} 
    
    & \begin{minipage}{0.3\linewidth} \centering 3 \includegraphics[width=\textwidth]{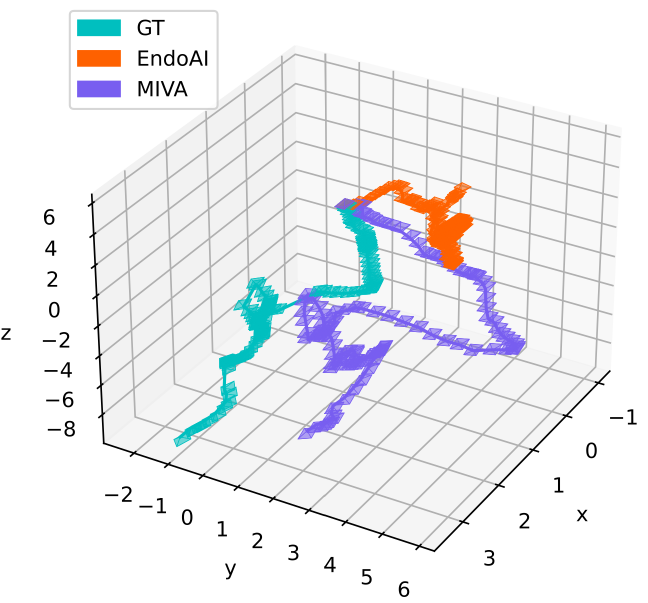} \\ \includegraphics[width=0.25\textwidth]{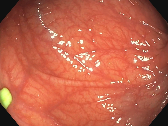} \includegraphics[width=0.25\textwidth]{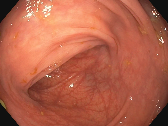} \includegraphics[width=0.25\textwidth]{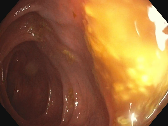}\end{minipage} 
    
    \\ \vspace{2em}

        \begin{minipage}{0.3\linewidth} \centering 4 \includegraphics[width=\textwidth]{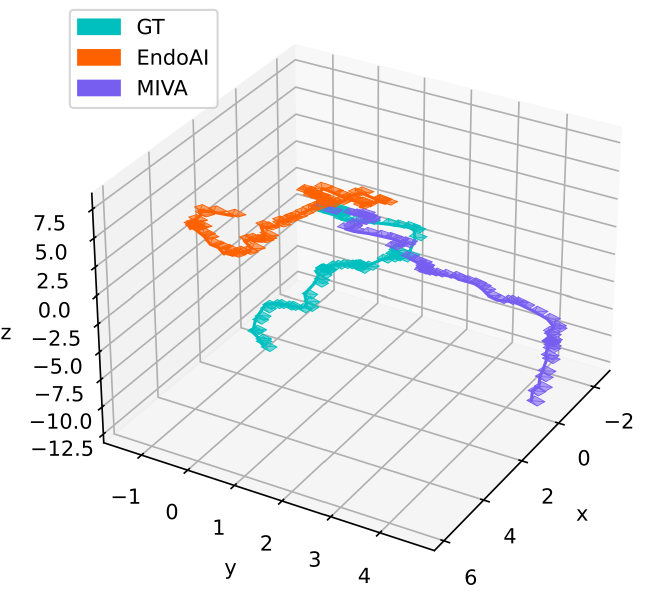} \\ \includegraphics[width=0.25\textwidth]{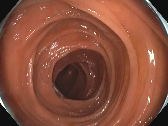} \includegraphics[width=0.25\textwidth]{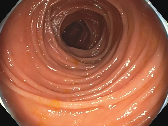} \includegraphics[width=0.25\textwidth]{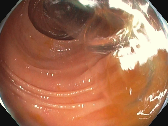}\end{minipage}
        
        &\begin{minipage}{0.3\linewidth} \centering 5 \includegraphics[width=\textwidth]{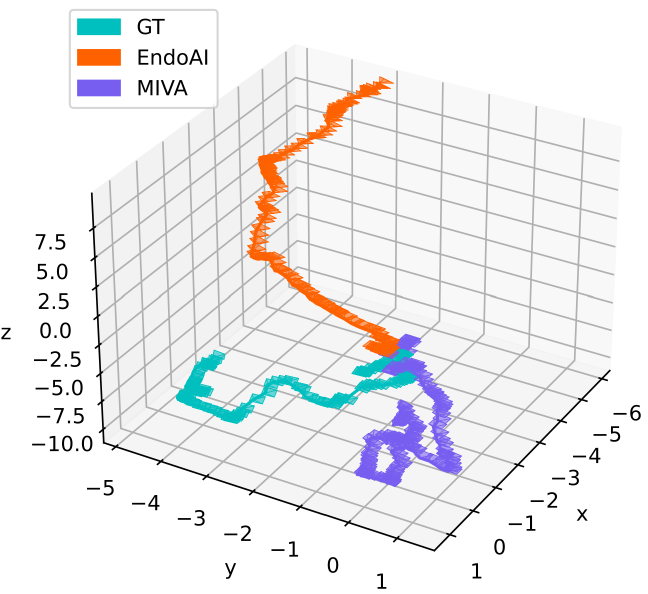} \\ \includegraphics[width=0.25\textwidth]{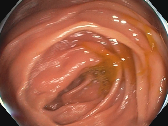} \includegraphics[width=0.25\textwidth]{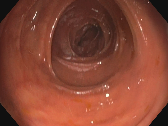} \includegraphics[width=0.25\textwidth]{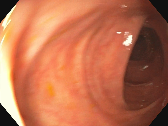}\end{minipage} 
        
        & \begin{minipage}{0.3\linewidth} \centering 6 \includegraphics[width=\textwidth]{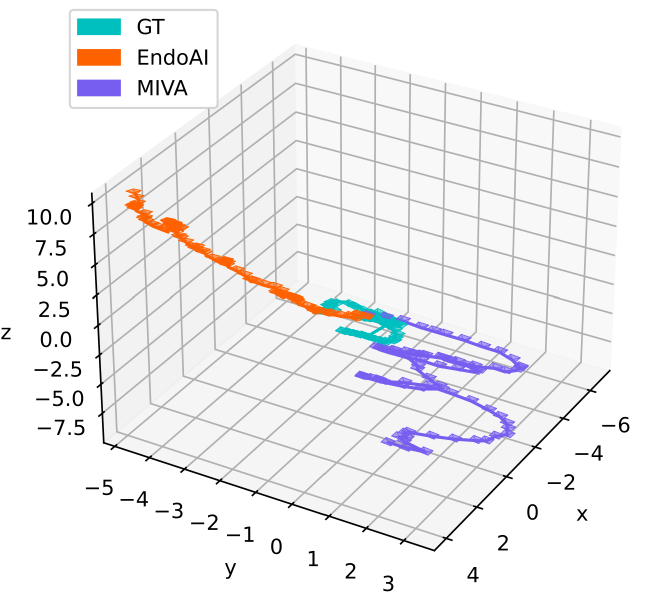} \\ \includegraphics[width=0.25\textwidth]{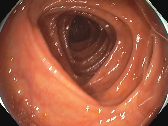} \includegraphics[width=0.25\textwidth]{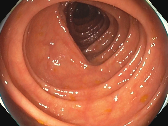} \includegraphics[width=0.25\textwidth]{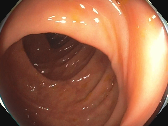}\end{minipage} \\

        &
            \begin{minipage}{0.3\linewidth} \centering 7 \includegraphics[width=\textwidth]{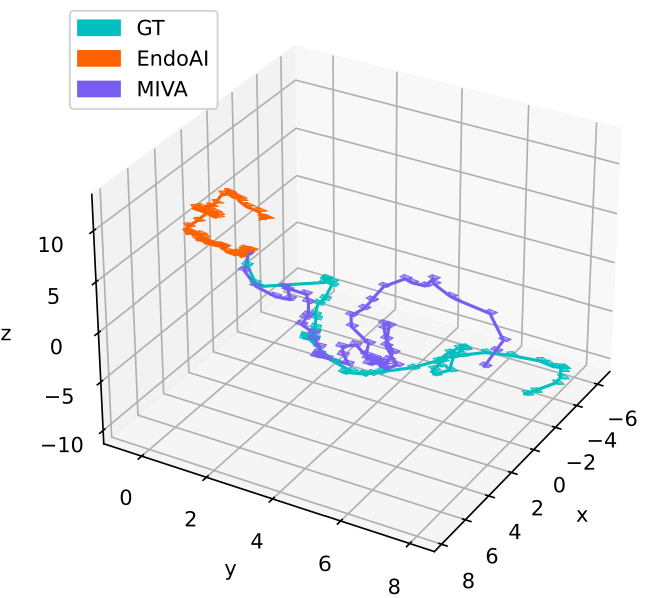} \\ \includegraphics[width=0.25\textwidth]{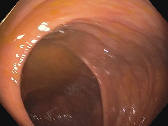} \includegraphics[width=0.25\textwidth]{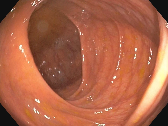} \includegraphics[width=0.25\textwidth]{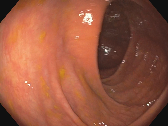}\end{minipage} &\\

    \end{tabular}
    \caption{Overview of Task 3 results showing predicted and ground truth (COLMAP) trajectories. For each test trajectory, we show three sampled frames in order of the video.}
    \label{tab:traj_t3}
\end{figure*}

\subsection{Task 3: Camera pose estimation on real-world data}

Two teams participated in Task 3. Both teams used the same method they also used for Task 2, but also applied a CycleGAN to translate appearance between the real and synthetic domains before predicting the camera pose.
Results are summarized in Table \ref{tab:res_task3}. As methods are compared to COLMAP, and overall scales are not known, the predicted trajectories are scaled before the evaluation of the error metrics. For the same reason, we are more interested in the ATE which better reflects the global consistency than the RTE. MIVA outperforms EndoAI on five out of seven scenes according to the ATE and is thus declared winner. MIVA also yields the smallest RTE in five of seven scenes. 

The methods are compared qualitatively in Figure \ref{tab:traj_t3}.   While both methods demonstrate extreme drift, the overall trajectories follow the COLMAP trajectory in some scenes, such as in scene 2, where both methods predict the sharp sideways movement in the first half of the trajectory. Similarly, both models show the quick sideways slip of the camera in the middle of the trajectory in scene 3. And in scene 5, both models follow the "W" shape of the trajectory.   We found that none of the participating groups used the publicly available COLMAP poses in the EndoMapper dataset for training. The power of their methods is based entirely on their pose models pretrained on synthetic data.   We thus posit that the synthetic data in this challenge provided the models with some understanding of camera pose movement in real colonoscopy.   Interestingly, the ranking of both teams is swapped in comparison to Task 2, although both teams use the same pose networks as before, and employ the same CycleGAN for domain adaptation.

\subsection{Data limitations and future directions}
  While synthetic data provides a useful playground to develop algorithms, its applicability to real procedures remains to be elucidated. While we strongly believe that synthetic datasets played a crucial role in enabling early research in the field \citep{mahmood2018deep} and in helping push the boundaries further \citep{rau2019implicit,mathew2020augmenting,itoh2021unsupervised,rodriguez2022uncertain}, drawbacks remain. First, the visual discrepancy between real images and our synthetically generated frames is obvious. But visual differences alone can usually be overcome with domain adaptation. More importantly, our synthetic data also misses some physical properties of real colons. For instance, colonoscopists often use water to clean the colon mucosa, resulting in puddles. Specularities and air bubbles are also common in real colonoscopies but are not reflected in our data. For the camera pose dataset, one important difference is the lack of deformation in the synthetic data. In the synthetic dataset, the movement of the colon wall is always due to a camera movement. But in real colonoscopy, the colon walls constantly move due to the colon's own digestive motions, or inflation with air.

So, while synthetic data is useful, the question of how we can move past having to choose between unrealistic synthetic data \textit{or} unlabeled real data remains unanswered. One obvious approach is improving the fidelity of synthetic data to replicate real colon mucosa more closely \citep{dowrick2023procedurally}. However, a domain gap is unavoidable, especially with respect to the behavior of the camera and the relative movement of the colon wall and haustral folds.   

A different approach is method-based and focuses on combining both modalities in a useful way \cite{rau2023task}. But to evaluate such methods, a real labeled dataset is indispensable. 

COLMAP provided useful ground truth poses for this challenge, but the method has serious limitations. It requires reliably matchable features which are extremely sparse in the colon. The reconstruction thus fails on many subsections of the colon. Even if it works, the resulting depth maps are too sparse to be useful, and depths and poses are biased toward a few visible features while ignoring most of the remaining colon wall. Due to the high failure rate, the reconstructions must be visually verified, further biasing the resulting test set towards sub-scenes that are visually interpretable. However, when COLMAP succeeds, it is accurate. We ran COLMAP on Synthetic Colon I and found that it fails to reconstruct 93\% of all frames but achieves an RTE of $0.028$ cm on the sections where it does not fail. For comparison, the best submission achieved $0.081$ cm on the entire Synthetic Colon I.

An alternative route for labeled real datasets could be new hardware. Magnetically actuated soft capsule endoscopes can provide partial ground truth pose, but not depth \citep{pittiglio2019magnetic}. Some capsule colonoscopes provide stereo vision, paving the door for more accurate, but still sparse, depth prediction \citep{bianchi2017innovative}. Similarly, full spectrum colonoscopy provides two additional lateral cameras \citep{kurniawan2017flexible}. While these advances currently focus on improving the visualization of the colonoscopic scenes for the operator in real-time, we hope that future advances incorporate other sensors, such as for position or depth.

A last alternative to synthetic data is colon phantoms made of synthetic materials, such as silicone. Phantoms are, perhaps, the most flexible approach. They can, in theory, be produced in any size, allowing the integration of mounted depth and pose sensors.  One drawback of phantoms is their material. Phantoms are either rigid, preventing a colonoscope from moving through it, especially around corners. Or they are non-rigid, rendering electromagnetic poses invalid as the sensor can move relative to the magnetic field while staying in place relative to the phantom. Further, the rubber-like surface looks unrealistic and prevents the camera from replicating realistic camera movements due to friction. As they are expensive to produce, a collection of many phantoms is unrealistic, such that data availability and diversity are limited. Lastly, hand-eye calibration between the camera and EM tracker and temporal synchronization introduces errors in the ground truth. Nonetheless, the creation of cheap and realistic looking and feeling phantoms could be a promising future direction.

\section{Conclusions}
This paper discusses the SimCol3D 2022 EndoVis Subchallenge and the methods employed by participating teams. The primary objective of this challenge was to promote research on 3D reconstruction during colonoscopy. Six teams from various parts of the world participated in the challenge and achieved impressive results. Particularly, the task of depth prediction on synthetic data proved to be both interesting and solvable. Achieving sub-millimeter accuracy on an unseen colon, the winning team could predict local 3D geometry extremely accurately. This robust generalization to a new scene within the same domain is a promising step towards real-world applications. The generalizability to a new domain remains an open research question and has not been addressed for the depth prediction task in this challenge. To test the applicability to real colonoscopy, new hardware facilitating datasets consisting of real colonoscopy frames with corresponding ground truth depth is required. While synthetic, phantom, and Structure-from-Motion-based data sources all have their own limitations, a thorough evaluation on all three modalities could paint a more holistic picture of model performance in the meantime. 

In comparison to depth prediction, predicting pose is a less well-studied problem, and accordingly the task is not yet fully solved. One main concern remains drift, which could be addressed by future work. Interestingly, both depth loss ($L_1$) and pose loss (RTE) increase roughly three-fold between the know scenes (I and II) versus the unseen scene (III). Therefore, it is crucial for future work to delve deeper into investigating the generalizability of models across different scenes, both within the same domain and across domains. While this challenge was the first one to evaluate generalizability from synthetic pose prediction to real procedures, the evaluation is limited by the quality of the COLMAP labels and their visual verification.

To have an impact on patient outcomes, accurate depth and pose predictions are a first step. Future work should tackle the challenge of achieving robust global reconstructions from local pose and depth predictions based on which unscreened colon mucosa can be identified and visualized. Such a framework will have to work in real-time and should be seamlessly integrate into clinical practice.

\section*{Acknowledgments}
The authors would like to thank the EndoVis team for facilitating this challenge and providing guidance and support. 

This work was supported by the Wellcome/EPSRC Centre for Interventional and Surgical Sciences (WEISS) [203145Z/16/Z]; Engineering and Physical Sciences Research Council (EPSRC) [EP/P027938/1, EP/R004080/1, EP/P012841/1]; The Royal Academy of Engineering Chair in Emerging Technologies scheme; and the EndoMapper project by Horizon 2020 FET (GA 863146). At the time of the challenge, the three corresponding authors were affiliated with University College London. 

The contribution of team CVML from the University of Central Lancashire was supported by the Science and Technology Facilities Council [ST/S005404/1].

For the purpose of open access, the author has applied a CC BY public copyright license to any author-accepted manuscript version arising from this submission.

The synthetic data supporting this work is openly available under a CC BY license at \url{https://www.ucl.ac.uk/interventional-surgical-sciences/simcol3d-data}.

Anita Rau, Sophia Bano, Yueming Jin, and Danail Stoyanov organized this challenge. Pablo Azagra, Javier Morlana, and Jos\'e M.M. Montiel curated the real dataset and generated COLMAP labels. Rawen Kader and Laurence B. Lovat clinically motivated and validated the project. All other authors were participants in the challenge.  All co-authors helped write this manuscript. 

\bibliographystyle{model2-names.bst}\biboptions{authoryear}
\bibliography{refs.bib}

\end{document}